\documentclass[mnsc,nonblindrev]{informs3_nojournal}

\usepackage{tikz,mathrsfs}
\usepackage{pgf}
\usepackage{graphicx}
\usetikzlibrary{arrows,automata}
\usepackage{amsfonts,amssymb}
\tikzstyle{legend}=[rectangle, draw=black,fill=white, rounded corners, minimum width=1cm, minimum height=0.75cm]
\OneAndAHalfSpacedXI
\usepackage{endnotes}
\usepackage{ mathrsfs }
\usepackage{graphicx}

\usepackage{bm}
\usepackage{algorithm2e}
\SetKwFor{For}{for (}{) $\lbrace$}{$\rbrace$}
\RestyleAlgo{boxruled}
 
\usepackage{eurosym}
\usepackage{hyperref}
\usepackage{enumitem}

\usepackage{multibib}
\newcites{RR}{Response References}
\usepackage[capitalise]{cleveref}

\let\footnote=\endnote

\newcommand{\E} { \mathbb{E} }

\newcommand{\conv}{\text{conv}}

\newcommand{\bbR}{\mathbb{R}}

\newcommand{\bbE}{\mathbb{E}}
\newcommand{\bbP}{\mathbb{P}}

\newcommand{\edit}{\textcolor{black}}

\newcommand{\calX}{\mathcal{X}}
\newcommand{\calH}{\mathcal{H}}
\newcommand{\calB}{\mathcal{B}}

\newcommand{\calC}{\mathcal{C}}
\newcommand{\calD}{\mathcal{D}}

\newcommand{\calS}{\mathcal{S}}
\newcommand{\calK}{\mathcal{K}}

\newcommand{\ellT}{\ell_{\mathrm{SPO}}}

\newcommand{\frakR}{\mathfrak{R}}

\newcommand{\diam}{\Delta}
\newcommand{\ldiam}{\omega_S}

\newcommand{\RadSPO}{\frakR^n_{\mathrm{SPO}}}
\newcommand{\RadHatSPO}{\hat\frakR^n_{\mathrm{SPO}}}
\newcommand{\ellTmar}{\ell_{\mathrm{SPO}}^\gamma}

\newcommand{\riskT}{R_{\mathrm{SPO}}}
\newcommand{\empRiskT}{\hat{R}_{\mathrm{SPO}}}

\newcommand{\empRiskTmar}{\hat{R}_{\mathrm{SPO}}^\gamma}
\newcommand{\RadHatSPOmar}{\hat\frakR^{n}_{\gamma\mathrm{SPO}}}
\newcommand{\empRiskTmark}{\hat{R}_{\mathrm{SPO}}^{\gamma_k}}
\newcommand{\riskTmar}{R_{\mathrm{SPO}}^\gamma}

\newcommand\SPOmar[1]{\ell_{\mathrm{mSPO}}^\gamma(#1, c)}

\newcommand{\Rad}{\frakR^n}
\newcommand{\RadHat}{\hat\frakR^n}

\newcommand\vsigma{\bm{\sigma}}

\newcommand\tr[1]{\mathrm{Tr}\left(#1\right)}

\newcommand{\dist}{\mathrm{dist}}
\newcommand{\Int}{\mathrm{int}}

\usepackage{natbib}
 \bibpunct[, ]{(}{)}{,}{a}{}{,}%
 \def\bibfont{\small}%
\TheoremsNumberedThrough     
\ECRepeatTheorems

\EquationsNumberedThrough    

\MANUSCRIPTNO{}

\begin{document}

\RUNAUTHOR{El Balghiti, Elmachtoub, Grigas, and Tewari}
\RUNTITLE{Generalization Bounds in the Predict-then-Optimize Framework}
\TITLE{Generalization Bounds in the \\ Predict-then-Optimize Framework}

\ARTICLEAUTHORS{%
\AUTHOR{Othman El Balghiti}
\AFF{  Department of Industrial Engineering and Operations Research, Columbia University, New York, NY, 10027,  \EMAIL{oe2161@columbia.edu}}
\AUTHOR{Adam N. Elmachtoub}
\AFF{Department of Industrial Engineering and Operations Research and Data Science Institute, Columbia University, New York, NY, 10027, \EMAIL{adam@ieor.columbia.edu}}
\AUTHOR{Paul Grigas}
\AFF{Department of Industrial Engineering and Operations Research, UC Berkeley, Berkeley, CA, 94720,  \EMAIL{pgrigas@berkeley.edu}}
\AUTHOR{Ambuj Tewari}
\AFF{Department of Statistics, University of Michigan, Ann Arbor, MI, 48109,  \EMAIL{tewaria@umich.edu}}
} 
\ABSTRACT{
The predict-then-optimize framework is fundamental in many practical settings: predict the unknown parameters of an optimization problem, and then solve the problem using the predicted values of the parameters. A natural loss function in this environment is to consider the cost of the decisions induced by the predicted parameters, in contrast to the prediction error of the parameters. This loss function was recently introduced in \cite{elmachtoub2017smart} and referred to as the Smart Predict-then-Optimize (SPO) loss. In this work, we seek to provide bounds on how well the performance of a prediction model fit on training data generalizes out-of-sample, in the context of the SPO loss. Since the SPO loss is non-convex and non-Lipschitz, standard results for deriving generalization bounds do not apply. 

We first derive bounds based on the Natarajan dimension that, in the case of a polyhedral feasible region, scale at most logarithmically in the number of extreme points, but, in the case of a general convex feasible region, have linear dependence on the decision dimension.
By exploiting the structure of the SPO loss function and a key property of the feasible region, which we denote as the strength property, we can dramatically improve the dependence on the decision and feature dimensions. Our approach and analysis rely on placing a margin around problematic predictions that do not yield unique optimal solutions, and then providing generalization bounds in the context of a modified margin SPO loss function that is Lipschitz continuous. Finally, we characterize the strength property and show that the modified SPO loss can be computed efficiently for both strongly convex bodies and polytopes with an explicit extreme point representation.
 
}

\maketitle

\section{Introduction}
A common application of machine learning is to \emph{predict-then-optimize}, i.e., predict unknown parameters of an optimization problem and then solve the optimization problem using the predictions. For instance, consider a navigation task that requires solving a shortest path problem. The key inputs into this problem are the travel times on each edge, typically called edge costs. Although the exact costs are not known at the time the problem is solved, the edge costs are predicted using a machine learning model trained on training data consisting of features (time of day, weather, etc.) and edge costs (collected from app data). Fundamentally, a good prediction model induces the optimization problem to find good shortest paths, as measured by the true edge costs. In fact, recent work has been developed to consider how to solve problems in similar environments \citep{bertsimas2014predictive,kao2009directed,donti2017task}. In particular, \cite{elmachtoub2017smart} developed the Smart Predict-then-Optimize (SPO) loss function for the case of optimization with an unknown linear objective subject to a convex feasible region, also known as contextual linear optimization. The SPO loss exactly measures the quality of a prediction by the decision error, in contrast to the prediction error as measured by standard loss functions such as mean squared error. A key property of a loss function is to show that the generalization error is small, i.e., that the loss observed in-sample is close to what we would observe out-of-sample. A generalization bound is an upper bound on the generalization error that holds with high probability, and should be relatively small to justify the use of minimizing the empirical risk on the training data. A generalization bound also implies a convergence rate on the number of samples required to find the best prediction model within a given hypothesis class. In this work, we seek to provide an assortment of novel and strong generalization bounds for the SPO loss function. 

Specifically, we shall assume that our optimization task is to minimize a linear objective over a feasible region, and that this task can be completed efficiently in the practical sense. This captures a wide range of optimization problems including linear programming, mixed-integer linear programming, and a large class of nonlinear programs. In the shortest path example, the feasible region is a polyhedron. We assume the objective cost vector is not known at the time that the optimization problem is solved, but rather predicted from a prediction model that maps from features to cost vectors. A decision is made with respect to the predicted cost vector, and the SPO loss is computed by evaluating the decision on the true cost vector and then subtracting the optimal cost assuming knowledge of the true cost vector. Unfortunately, the SPO loss is non-convex and non-Lipschitz, and therefore proving generalization bounds is not immediate.

Our results consider two approaches, and in all cases achieve bounds that are $\tilde{O}(\frac{1}{\sqrt{n}})$, where $n$ is the number of samples. Each approach uses the classic result of \cite{bartlett2002rademacher} (Theorem \ref{SPO_rademacher}), which allows us to bound the generalization error by using the Rademacher complexity of the hypothesis class composed with predict-then-optimize models. In the first and more straightforward approach, we treat the optimization task as a generalization of multiclass classification (see Example \ref{ex:spo}). When the feasible region is a polyhedron, this resorts to treating every extreme point as a class. The final bound depends on the square root of the Natarajan dimension of the hypothesis class, and the square root of the log of the number of extreme points (Theorem \ref{Main_Theo}). The Natarajan dimension essentially measures how rich the hypothesis class of prediction models is, and in the case of linear models it is exactly the number of parameters $pd$, where $p$ is the feature dimension and $d$ is the decision dimension (Corollary \ref{cor:linear}). For a general convex body, we transform the optimization task into a multiclass classification problem by discretizing the feasible region with a ball covering that only results in a loss of $O(\frac{1}{n})$ (Corollary \ref{cor:balls}). The ball covering can be exponential in $d$, and the final generalization bound can thus depend on $d\sqrt{p}$ in the case of a linear hypothesis class. 


Fortunately, we show that for a large class of feasible regions, tighter generalization bounds can be obtained  with improved dependence on $p$ and $d$, using margin-based methods. The key idea of our approach is to reduce the analysis of the Rademacher complexity of the SPO framework to a multivariate Rademacher complexity of the vector-valued hypothesis class, for which we can obtain improved bounds in many cases. This reduction relies on a vector contraction inequality that only works for Lipschitz loss functions. Since SPO loss is not Lipschitz, we transform it to a Lipschitz loss function by using a margin-based analysis that bounds away predictions that yield multiple optimal solutions (and cause discontinuities in the SPO loss). We define a property of the feasible region which we call the strength property, and show that our margin-based loss function is Lipschitz if the strength property is satisfied (Theorem \ref{thm:main_lipschitz}). In turn, this allows to provide generalization bounds with respect to the margin-based loss, the strength parameter, and with the more amenable Rademacher complexity of the hypothesis class (Theorem \ref{thm:margin_generalization}). We also show the margin parameter can be cross-validated by showing the generalization bound holds uniformly over an interval of margin parameters (Theorem \ref{thm:margin_generalization_uniform}). We show that for suitably constrained linear hypothesis classes, we get a much improved dependence on problem dimensions (Theorem \ref{thm:linear_class}). Finally, we show the strength property and how to compute the  margin-based loss for strongly convex sets (Theorem \ref{thm:strongly}) and polyhedra with an explicit extreme point representation (Theorem \ref{thm:poly}).

\subsection{Relation to Previous Work}



We first summarize a few additional approaches proposed in the literature which for solving predict-then-optimize problems. \cite{kao2009directed} propose a loss function for training linear regression models which minimizes a convex combination between the prediction error and decision error of an unconstrained quadratic optimization.  \cite{donti2017task} provide a more general methodology that allows for constraints which relies on differentiating the optimization problem using a technique from \cite{amos2017optnet}.  
\cite{wilder2019melding, mandi2019smart,ferber2020mipaal, wilder2019end, poganvcic2019differentiation, mandi2020interior,berthet2020learning} consider training ML models using ``decision-focused'' loss functions for various linear and combinatorial optimization problems; their methods do not attempt to minimize SPO loss directly but rather employ differentiable surrogate loss functions. \cite{elmachtoub2017smart} propose a convex surrogate loss funcion called SPO+ loss which is shown to be Fisher consistent with SPO loss.  \cite{loke2021decision} consider a form of regularization that generalizes the SPO+ loss function. 
\citet{demirovic2019predict+, demirovic2020dynamic} propose algorithms for training linear regression models to directly minimize SPO loss, but their methodology is specialized for ranking optimization problems and dynamic programming problems, respectively.  \cite{ho2020risk} study the relationship between the mean squared error loss and the SPO loss. See \cite{kotary2021end} for a recent survey of more approaches. The SPO loss has been used in applications such as last mile delivery \citep{chu2021data} and ship inspection \citep{yan2020semi}. \cite{elmachtoub2020decision} and \cite{kallus2020stochastic} train decision trees and random forests to directly minimize decision error in contextual linear and nonlinear optimization problems, respectively. 

 A preliminary version of this paper appeared in \cite{el2019generalization} that did not include our general framework for margin-based bounds using the strength property. Our results have found use in \cite{hu2020fast} which slightly improved Theorem \ref{Main_Theo} by a factor of $\sqrt{\log n}$. They also show - in settings where there is no model misspecification (realizability) - how to obtain regret rates that can be faster than $O(\frac{1}{\sqrt{n}})$ when using SPO loss and least squares loss. \cite{wang2020automatically} uses our Theorem \ref{Main_Theo} to derive generalization bounds on compact surrogate loss functions. Generalization bounds for related types of  decision-making problems have been derived in  \cite{ban2019big, ho2019data, bertsimas2014predictive}.

As previously mentioned, the SPO framework is a generalization of multi-class classification, and thus our results have some connections to results in this literature stream. In fact, our results can be seen as extending the classic margin-based generalization bounds for classification \citep{koltchinskii2002empirical, koltchinskii2001some, mohri2018foundations} to the predict-then-optimize framework. Note that the reduction of a predict-then-optimize problem to multi-class classification throws away potentially important information, namely the numerical values of the cost vectors. Therefore a naive application of results in the classification literature would not lead to bounds on our loss function of interest, the SPO loss. 
Furthermore, our approaches remove a strong explicit dependency on the number of extreme points (classes) by exploiting the structure of the SPO loss. Our data-independent bounds grow only logarithmically in the number of extreme points. 
In contrast, for data-independent worst-case bounds in multiclass classification, the dependency is at best square root in the number of classes \citep{guermeur2007vc,daniely2015multiclass}. Our data-dependent margin based bounds remove an explicit dependency on the number of extreme points entirely.
Since the seminal work of \cite{koltchinskii2002empirical}, much work in the classification literature has focused on improving the dependency on the number of classes for nonparameteric hypothesis classes such as multi-class kernels.  
For example, by using data-dependent (margin-based) approaches and local Rademacher complexity analysis, \cite{lei2015multi} and \cite{NIPS2018_7431} obtain logarithmic complexity in the number of classes for such hypothesis classes.

\section{Predict-then-Optimize Framework} \label{sec:framework}
We now formally describe the predict-then-optimize framework, which is central to many applications of machine learning for use in operations research problems. We are given a nominal optimization problem of interest which models a downstream decision-making task. We assume that the nominal problem has a linear objective and that the decision variable $w \in \bbR^d$ and feasible region $S \subseteq \bbR^d$ are well-defined and known with certainty. We assume $S$ is a nonempty, compact, and convex set.
However, the cost vector of the objective, $c \in \bbR^d$, is not observed directly, and rather an associated feature vector $x \in \bbR^p$ is observed. Let $\calD$ be the underlying joint distribution of $(x, c)$ and let $\mathcal{D}_x$ be the conditional distribution of $c$ given $x$. Then, the goal for the decision maker is to solve the contextual stochastic optimization problem
\begin{align}
\min_{w \in S} \E_{c \sim \mathcal{D}_x}[c^T w |x] \ = \ \min_{w \in S} \E_{c \sim \mathcal{D}_x}[c|x]^T w \ . \label{core}
\end{align}
The predict-then-optimize framework relies on using a prediction (estimate) for $\E_{c \sim \mathcal{D}_x}[c|x]$, which we denote by $\hat{c}$, and solving the deterministic version of the optimization problem based on $\hat{c}$. From \eqref{core}, it is clear that the linearity of the objective makes the predict-then-optimize approach viable, where as a nonlinear objective may require an alternate approach such as that developed by \cite{bertsimas2014predictive}. Note that knowledge of the distribution $\mathcal{D}_x$ is typically unavailable, thus necessitating a data-driven approach to solving \eqref{core}. 

We define $P(\hat{c})$ to be the optimization task with objective cost vector $\hat{c}$, namely 
\begin{equation}\label{poi}
\begin{array}{rcl}
P(\hat{c}):~ & \min\limits_{w} & \hat{c}^T w \\
& \text{s.t.} & w \in S. \
\end{array} 
\end{equation}
 Since we are focusing on linear optimization problems, the assumption that $S$ is convex is without loss of generality as we can take $S$ to be the convex hull of any potentially non-convex feasible region.
 We let $w^\ast(\cdot) : \bbR^d \to S$ denote any oracle for solving $P(\cdot)$. That is, $w^\ast(\cdot)$ is a fixed deterministic mapping such that $w^\ast(c) \in \arg\min_{w \in S}\left\{c^T w\right\}$ for all $c \in \bbR^d$. For instance, if \eqref{poi} corresponds to a linear, conic, or mixed-integer optimization problem (in which case $S$ can be implicitly described as a convex set), then a commercial optimization solver or a specialized algorithm suffices for $w^\ast(\cdot)$.

In this framework, we assume that predictions are made from a model that is learned on a training data set. Specifically, we are provided sample training data  $(x_1, c_1), \ldots, (x_n, c_n)$ drawn i.i.d.\ from the joint distribution $\mathcal{D}$, where $x_i \in \calX \subseteq \mathbb{R}^p$ is a feature vector representing auxiliary information associated with the cost vector $c_i \in \calC \subseteq \mathbb{R}^d$. 
We denote by $\calH$ our hypothesis class of cost vector prediction models. Thus, for a function $f\in \calH$, we have that $f: \calX \to \bbR^d$. Most approaches for learning a model $f \in \calH$ from the training data are based on specifying a loss function that quantifies the error in making prediction $\hat c$ when the realized (true) cost vector is actually $c$.
Following prior work of \cite{elmachtoub2017smart}, our primary loss function of interest is the Smart Predict-then-Optimize (SPO) loss function that directly takes the nominal optimization problem $P(\cdot)$ into account when measuring errors in predictions. Namely, we consider the SPO loss function defined by:
\begin{equation*}
\ellT(\hat{c}, c) := c^T w^\ast(\hat{c}) - c^T w^\ast(c)  \ ,
\end{equation*}
where $\hat{c}$ is the predicted cost vector and $c$ is the true realized cost vector. The SPO loss can be described as the decision error of making the decision based on $\hat{c}$ when the actual cost vector was $c$, evaluated on the actual cost vector $c$. The SPO loss is always non-negative and is exactly zero when the induced decisions from $\hat{c}$ and $c$ are equal, even if $\hat{c}$ and $c$ themselves are not necessarily close (which is traditionally the goal in prediction). 
\edit{Note that, primarily for simplicity of presentation, we consider the version of the SPO loss defined relative to the specific and fixed optimization oracle $w^\ast(\cdot)$ (Definition 1 of \cite{elmachtoub2017smart}) rather than the unambiguous variant that does not depend on a specific optimization oracle (Definition 2 of \cite{elmachtoub2017smart}).}

Given a fixed sample $(x_1,c_1),\ldots,(x_n,c_n)$, we define the empirical risk with respect to the SPO loss of a function $f \in \calH$ as
\[
\empRiskT(f) := \frac{1}{n} \sum_{i=1}^n \ellT(f(x_i), c_i) \ ,
\]
and the expected (or Bayes) risk as $$\riskT(f) := \bbE_{(x,c)\sim \calD}[\ellT(f(x),c)].$$ One can interpret the expected risk as the out-of-sample performance of $f$, and the empirical risk as the in-sample performance. The empirical risk minimization principle states that we should determine a prediction model $\hat{f}_n \in \calH$ by solving the optimization problem 
\begin{equation}\label{erm1}
\min_{f \in \calH} \ \empRiskT(f)  \ . 
\end{equation}
\edit{Empirical risk minimization with respect to the SPO loss is not the only way to train a prediction model.
Indeed, as pointed out in \cite{elmachtoub2017smart}, the SPO loss function is generally non-convex, may even be discontinuous, and is in fact a strict generalization of the 0-1 loss function in binary classification. Thus, solving \eqref{erm1} to optimally is intractable even when $\calH$ is a linear hypothesis class. To circumvent these difficulties, alternative approaches include optimizing a convex surrogate loss such as the SPO+ loss presented in \citep{elmachtoub2017smart} or the squared $\ell_2$ loss, using methods exploiting structural properties of $\calH$ \citep{elmachtoub2020decision}, or modifying $P(\cdot)$ and finding local optimal solutions \citep{wilder2019melding}. 
Most generally, one may use any training procedure that determines a model in the hypothesis class $\calH$ and therefore it is important that generalization bounds apply \emph{uniformly} for all functions in $\calH$.
Our focus herein is on deriving generalization bounds that hold uniformly over a given hypothesis class $\calH$ and thus are valid for \emph{any} training approach, including the use of a surrogate or alternate loss function within the framework of empirical risk minimization.
We also note that a generalization bound for the SPO loss directly translates to an upper bound guarantee for problem \eqref{core} that holds ``on average'' over the distribution.}

\edit{Provided with any prediction model $\hat{f}$ trained on the given data, we can now simply express the predict-then-optimize approach to solve \eqref{core}; when presented with a feature vector $x$, select the optimal solution with respect to the predicted cost vector, i.e., the decision made is $w^\ast(\hat{f}(x))$. Ideally, in order to select see if $\hat{f}$ is a good model, we would evaluate the expected future real-world performance of $\hat{f}$,  i.e., the decision-maker would evaluate the expected risk $\riskT(\hat{f})$. Unfortunately, as the true distribution is unknown, one can only estimate this quantity using data and the most natural estimate is the empirical risk $\empRiskT(\hat{f})$. Generalization bounds, which bound the gap between the empirical and expected risk with high probability, are thus critical as they validate that $\empRiskT(\hat{f})$ is a reasonable estimate of $\riskT(\hat{f})$. Strong validation bounds validate the usage of empirical risk minimization and other model selection ideas that estimate the empirical risk on the training data.  Indeed, one may use a (two-sided) generalization bound to construct a confidence interval for $\riskT(\hat{f})$, for example. At a more cursory level, generalization bounds reveal how the difference between the expected and empirical risks scale with key parameters such as the sample size and dimensions of the feature and decisions spaces.}

To conclude this subsection, we present several important examples to illustrate the applicability and generality of the SPO loss function and framework.

\begin{example}[Shortest Path]
In the shortest path problem, the feature vector $x$ may include features such as weather, speed limit, and time data that may be used to predict the cost vector $c$ representing the travel times along each edge of the network. In this case, the network is assumed to be given (e.g., the road network of a city) and the feasible region $S$ is a network flow polytope that represents flow conservation and capacity constraints on the underlying network. More generally, the class of vehicle routing problems with unknown edge costs falls in our framework, although the feasible region is more complex as integer constraints are typically required. 
\Halmos \end{example}

\begin{example}[Portfolio Optimization]
In portfolio optimization, the returns of potential investments can depend on many features which typically include historical returns, news, economic factors, social media, and others. We presume that these  features may be used to predict the vector of returns different assets, but that the covariance matrix of the asset returns does not depend on the features.  Maximizing the expected return is a linear objective, and with convex constraints that the total amount invested is in the simplex and the overall variance of the portfolio is bounded. 
\Halmos \end{example}

\begin{example}[Multiclass Classification] \label{ex:spo}
Our setting also captures multi-class (and binary) classification by the following characterization: $S$ is the $d$-dimensional unit simplex, namely $S := \{w \in \bbR^d : \sum_{j = 1}^d w_j = 1, w \geq 0\}$, where $d$ is the number of classes, and $\calC=\{-e_i| i=1,\ldots,d\}$ where $e_i$ is the $i^{th}$ unit vector in $\mathbb{R}^d$. It is easy to see that each vertex of the simplex corresponds to a class, and correct/incorrect classification has a loss of 0/1. Specifically, $\ellT(\hat{c},c)=0$ if $w^*(\hat{c})=-c$ and $\ellT(\hat{c},c)=1$ if $w^*(\hat{c})\neq -c.$
\Halmos \end{example}




\subsection{Other Notation}
We make use of a generic given norm $\|\cdot\|$ on $w \in \bbR^d$, as well as the $\ell_q$-norm denoted by $\|\cdot\|_q$ for $q \in [1, \infty]$. For the given norm $\|\cdot\|$ on $\mathbb{R}^d$, $\|\cdot\|_\ast$ denotes the dual norm defined by $\|c\|_\ast := \max_{w : \|w\| \leq 1} c^Tw$. Let $B(\bar{w}, r) := \{w : \|w - \bar{w}\| \leq r\}$ denote the ball of radius $r$ centered at $\bar{w}$, and we analogously define $B_q(\bar{w}, r)$ for the $\ell_q$-norm and $B_{\ast}(c, r)$ for the dual norm. 
For a set $W \subseteq \bbR^d$, let $\dist_W(\bar{w}) := \inf_{w \in W}\left\{\|w - \bar{w}\|\right\}$ be the distance from the point $\bar{w} \in \bbR^d$ to $W$ measured in the norm $\|\cdot\|$. Likewise, for a set $C \subseteq \bbR^d$, define $\dist_C^\ast(\bar{c}) := \inf_{c \in C}\left\{\|c - \bar{c}\|_\ast\right\}$ which is measured in the dual norm.
For a set $S \subseteq \bbR^d$, we define the size of $S$ in the norm $\|\cdot\|$ by $\rho(S) := \sup_{w \in S}\|w\|$, and the diameter of $S$ by $\diam(S) := \sup_{w_1, w_2 \in S}\|w_1 - w_2\|$. We analogously define $\rho_q(\cdot)$ for the $\ell_q$-norm and $\rho_\ast(\cdot)$ for the dual norm. We define the ``linear optimization gap'' of $S$ with respect to $c$ by $\ldiam(c) := \max_{w \in S}\left\{c^Tw\right\} -  \min_{w \in S}\left\{c^Tw\right\}$, and for a set $\calC \subseteq \bbR^d$ we slightly abuse notation by defining $\ldiam(\calC) := \sup_{c \in \calC}\ldiam(c)$. Note that the $\ellT(\hat{c},c)$ is non-negative and bounded above by $\ldiam(\calC)$ for all $\hat{c} \in \bbR^d$ and $c \in \calC$. Finally, we define the class of predict-then-optimize functions by $w^*(\mathcal{H}):= \{x \mapsto w^\ast(f(x)) : f \in \calH\}$. \\

\subsection{Rademacher Complexity and Generalization Bounds}

Let us now briefly review the notion of Rademacher complexity and its application in our framework.  We define the empirical Rademacher complexity of $\calH$ with respect to the SPO loss, i.e., the empirical Rademacher complexity of the function class obtained by composing $\ellT$ with $\calH$, by
\begin{equation*}
\RadHatSPO(\calH) := \bbE_{\sigma}\left[\sup_{f \in \calH}\frac{1}{n}\sum_{i = 1}^n \sigma_{i}\ellT(f(x_i), c_i)\right] \ ,
\end{equation*}
where $\sigma_{1}, \ldots, \sigma_n$ are i.i.d. Rademacher random variables, i.e., $\mathbb{P}(\sigma_i=1)=\mathbb{P}(\sigma_i=-1)=\frac{1}{2}$. The expected version of the Rademacher complexity is defined as $\RadSPO(\calH) := \bbE\left[\RadHatSPO(\calH)\right]$ where the expectation is w.r.t.\ an i.i.d.\ sample drawn from the underlying distribution $\calD$. This Rademacher complexity intuitively measures how expressive $\calH$ is in terms of correlating $\ellT(f(x_i),c_i)$ with random noise $\sigma_i$. The following theorem is an adaptation of the classical generalization bounds based on Rademacher complexity due to \cite{bartlett2002rademacher} to our setting.

\begin{theorem}[\cite{bartlett2002rademacher}, Theorem 3.3 in \cite{mohri2018foundations}] \label{SPO_rademacher}
Let $\calH$ be a family of functions mapping from $\calX$ to $\bbR^d$. Then, for any $\delta > 0$, with probability at least $1 - \delta$ over an i.i.d.\ sample drawn from the distribution $\calD$, each of the following holds for all $f \in \calH$:
\begin{align*}
\riskT(f) &\leq \empRiskT(f) + 2\RadSPO(\calH) + \ldiam(\calC)\sqrt{\frac{\log(1/\delta)}{2n}} \ .
\end{align*}
\end{theorem}

\edit{
\begin{remark}
Note that the above Theorem provides a one-sided upper generalization bound and that its proof relies on McDiarmid's inequality. A two-sided generalization bound can be obtained with only additional absolute constants by applying McDiarmid's inequality twice. Such a result yields upper and lower bounds on $\riskT(f)$, enabling the construction of a two-sided confidence interval. Since we are ultimately interested in minimizing $\riskT(f)$, our focus herein is on providing one-sided upper generalization bounds; however many of our results (with the notable exception of the margin-based bounds in Section \ref{sec:margin}) can readily be extended to two-sided bounds for $\riskT(f)$. \Halmos
\end{remark}
}

\edit{As previously noted, we are interested in guarantees that hold across a wide range of training procedures, beyond simply the empirical risk minimizer of the SPO loss. The fact that Theorem \ref{SPO_rademacher} holds \emph{uniformly} over the hypothesis class $\calH$ critically provides us with a tool for obtaining such guarantees:  as long as the training procedure returns a predictor in $f \in \calH$ and $\calH$ has a bounded Rademacher complexity $\RadSPO(\calH)$, then Theorem \ref{SPO_rademacher} yields a useful upper bound on the out-of-sample SPO risk $\riskT(f)$. We also note that, equipped with Theorem \ref{SPO_rademacher}, we obtain ``for free" an additional guarantee on the \emph{excess risk} of the empirical risk minimizer of the SPO loss $\hat{f}_n$. Indeed, Corollary \ref{SPO_rademacher_cor} below is a standard combination of Theorem \ref{SPO_rademacher} with Hoeffding's inequality. The full proof is in Appendix \ref{cor:SPO_rademacher_cor}.}

\edit{
\begin{corollary}\label{SPO_rademacher_cor}  Let $\calH$ be a family of functions mapping from $\calX$ to $\bbR^d$. Then, for any $\delta > 0$, with probability at least $1 - \delta$ over an i.i.d.\ sample drawn from the distribution $\calD$, any empirical risk minimizer $\hat{f}_n$ solving \eqref{erm1} satisfies:
\begin{align*}
\riskT(\hat{f}_n) - \min_{f \in \calH} \riskT(f) &\leq 2\RadSPO(\calH) + 2\ldiam(\calC)\sqrt{\frac{\log(2/\delta)}{2n}} \ .
\end{align*}
\end{corollary}
} 

Finally, we review an extension of Rademacher complexity to the case of vector-valued function classes.  Following \cite{bertsimas2014predictive} and \cite{maurer2016vector}, given a fixed sample $ (x_1,c_1),\ldots,(x_n,c_n)$, we define the {\em multivariate empirical Rademacher complexity} of $\calH$ as
\begin{equation}\label{rad_multi}
\RadHat(\calH) := \bbE_{\sigma}\left[\sup_{f \in \calH}\frac{1}{n}\sum_{i = 1}^n\sum_{j = 1}^d \sigma_{ij}f_j(x_i)\right] =
\bbE_{\vsigma}\left[\sup_{f \in \calH}\frac{1}{n}\sum_{i = 1}^n \vsigma_{i}^T f(x_i)\right] \ ,
\end{equation}
where $\sigma_{ij}$ are i.i.d.\ Rademacher random variables for $i = 1, \ldots, n$ and $j = 1, \ldots, d$, and $\vsigma_i := (\sigma_{i1},\ldots,\sigma_{id})^T$. The expected version of the multivariate Rademacher complexity is defined as $\Rad(\calH) := \bbE\left[\RadHat(\calH)\right]$ where the expectation is taken with respect to the i.i.d.\ sample drawn from the underlying distribution $\calD$. 

\section{Generalization Bounds Based on Multiclass Classification} \label{sec:polyhedron}

In this section, we primarily consider the case where $S$ is a polyhedron and derive generalization bounds based on bounding the Rademacher complexity of $\calH$ with respect to the SPO loss and applying Theorem \ref{SPO_rademacher}. The idea is similar to the bounds show for multiclass classification   based on Natarajan dimension \cite[Ch. 29]{shalev2014understanding}, where we now treat each extreme point as a class label.
Since $S$ is polyhedral, the optimal solution of $P(\cdot)$ can always be found by considering only the finite set of extreme points of $S$, which we denote by the set $\mathfrak{S}$.  Since the number of extreme points may be exponential in $d$, our goal is to provide bounds that are logarithmic in $|\mathfrak{S}|$. Note that, in the case of ties, \edit{we do not require} that $w^\ast(\hat{c})$ necessarily returns an extreme point of $S$.
At the end of the section, we extend our analysis to any compact and convex feasible region $S$ by extending the polyhedral analysis with a covering number argument. 

In order to derive a bound on the Rademacher complexity, we  critically rely on the notion of  Natarajan dimension \citep{natarajan1989learning}, which is an extension of the VC-dimension to the multiclass classification setting and is defined in our setting as follows.

\begin{definition}[Natarajan dimension]
Let $\mathcal{F} \subseteq S^{\mathcal{X}}$ be a hypothesis space of functions mapping from $\calX$ to $S$, and let $\mathbb{X} \subseteq \mathcal{X}$ be given. We say that $\mathcal{F}$ N-shatters $\mathbb{X}$ if there exists  $g_1, g_2: \mathbb{X} \rightarrow S$ such that 
\begin{itemize}
\item  $g_1(x)\not = g_2(x)$ for all $x \in \mathbb{X}$
\item For all $T \subseteq \mathbb{X}$, there exists $g \in \mathcal{F}$ such that (i) for all $x \in T \text{, }g(x)=g_1(x) $ and (ii) for all $x \in \mathbb{X} \backslash T \text{, }g(x)=g_2(x)$.
\end{itemize}
The Natarajan dimension of $\mathcal{F}$, denoted $d_N({\mathcal{F}})$, is the maximal cardinality of a set N-shattered by $\mathcal{F}$. \Halmos
\end{definition}

Notice that our definition of the Natarajan dimension above is a slight generalization of the standard definition, as we allow for functions that map to an infinite set $S$ instead of a finite set of ``labels." In our context, the labels correspond to the optimal decisions induced by the optimization oracle $w^\ast(\cdot)$. The slight generalization we consider allows us to accommodate cases when $w^\ast(\hat{c})$ is not necessarily an extreme point of $S$ (in the case of a tie).
The Natarajan dimension is a measure for the richness of a hypothesis class. In Theorem \ref{Main_Theo}, we show that the Rademacher complexity for the SPO loss can be bounded as a function of the Natarajan dimension of $w^*(\mathcal{H}):= \{x \mapsto w^\ast(f(x)) : f \in \calH\}$ and the number of extreme points of $S$. The proof, in Appendix \ref{sec:proofthm2}, follows a classical argument and makes strong use of Massart's lemma and the Natarajan lemma.

\begin{theorem}\label{Main_Theo} 
Suppose that $S$ is a polyhedron and $\mathfrak{S}$ is the set of its extreme points.
Let $\calH$ be a family of functions mapping from $\calX$ to $\bbR^d$. Then  we have that
\begin{align*} 
\RadSPO(\calH)\leq \ldiam(\calC)\sqrt{\frac{2d_N(w^*(\mathcal{H}))\log(n|\mathfrak{S}|^2)}{n}}.
\end{align*}
Furthermore, for any $\delta > 0$, with probability at least $1 - \delta$ over an i.i.d.\ sample $(x_1,c_1),\dots, (x_n,c_n)$ drawn from the distribution $\calD$, for all $f \in \calH$ we have
\begin{align*}
\riskT(f) &\leq \empRiskT(f) + 2\ldiam(\calC)\sqrt{\frac{2d_N(w^*(\mathcal{H}))\log(n|\mathfrak{S}|^2)}{n}} + \ldiam(\calC)\sqrt{\frac{\log(1/\delta)}{2n}} .
\end{align*} 
\end{theorem}

Next, we show that when $\calH$ is restricted to the linear hypothesis class $\calH_{\mathrm{lin}}=\{ x \mapsto Bx : B \in \mathbb{R}^{d \times p}\}$, then the Natarajan dimension of $w^*(\mathcal{H}_{lin})$ can be bounded by $dp$.
The proof, in Appendix \ref{sec:cor1proof},  relies on translating our problem to an instance of a linear multiclass prediction problem and using a result of \cite{daniely2014optimal}.
\begin{corollary} \label{cor:linear}
Suppose that $S$ is a polyhedron and $\mathfrak{S}$ is the set of its extreme points. Let $\mathcal{H}_{lin}$ be the hypothesis class of all linear functions, i.e., $\calH_{\mathrm{lin}}=\{ x \mapsto Bx : B \in \mathbb{R}^{d \times p}\}$.  Then we have \begin{align*}
d_N(w^*(\calH_{\mathrm{lin}}))\leq dp.
\end{align*}
Furthermore, for any $\delta > 0$, with probability at least $1 - \delta$ over an i.i.d.\ sample $(x_1,c_1),\dots, (x_n,c_n)$ drawn from the distribution $\calD$, for all $f \in \calH_{\mathrm{lin}}$ we have
\begin{align*}
\riskT(f) &\leq \empRiskT(f) + 2\ldiam(\calC)\sqrt{\frac{2dp \log(n|\mathfrak{S}|^2)}{n}} + \ldiam(\calC)\sqrt{\frac{\log(1/\delta)}{2n}}.
\end{align*} 
\end{corollary}

Next, we will build off the previous results to prove a generalization bound in the case where $S$ is a general compact convex set. The arguments we made earlier made extensive use of the extreme points of the polyhedron. Nevertheless, this combinatorial argument can be modified in order to derive similar results for general $S$. The approach is to approximate $S$ by a grid of points corresponding to the smallest cardinality $\epsilon$-covering of $S$. To optimize over these grid of points, we first find the optimal solution in $S$ and then round to the nearest point in the grid. Both the grid representation and the rounding procedure can fortunately both be handled by similar arguments made in Theorems \ref{Main_Theo} and Corollary \ref{cor:linear}, yielding a generalization bound below. The full proof is in Appendix \ref{sec:cor2proof}.

\begin{corollary} \label{cor:balls}
Let $S$ be any compact and convex set, and let $\calH_{\mathrm{lin}}$ be the hypothesis class of all linear functions.
Then, for any $\delta > 0$, with probability at least $1 - \delta$ over an i.i.d.\ sample $(x_1,c_1),\dots, (x_n,c_n)$ drawn from the distribution $\calD$, for all $f \in \calH_{\mathrm{lin}}$ we have
\begin{align*}
\riskT(f) &\leq \empRiskT(f) + 4d\ldiam(\calC) \sqrt{\frac{2p \log ( 2n\rho_2(S) d)}{n}} + \ldiam(\calC)\sqrt{\frac{\log(1/\delta)}{2n}} + O\left(\frac{1}{n}\right).
\end{align*}
\end{corollary}
Although the dependence on the sample size $n$ in the above bound is favorable, the dependence on the \edit{dimension of the feature space} $p$ and the dimension of the feasible region $d$ is relatively weak.  
Given that the proofs of Corollary \ref{cor:balls} and Theorem \ref{Main_Theo} are purely combinatorial and hold for worst-case distributions, this is not surprising. In the next section, we demonstrate how to exploit the structure of the SPO loss function and additional properties of the distribution $\calD$ and the feasible region $S$ in order to develop improved bounds.


\section{Margin-based Generalization Bounds}\label{sec:margin}
In this section, we seek to obtain generalization bounds that have better dependency on the feature dimension $p$ and the decision dimension $d$ than those in Corollaries \ref{cor:linear} and \ref{cor:balls}. The key idea in Section \ref{sec:polyhedron} is to essentially treat $S$ as a collection of extreme points and treat the predict-then-optimize problem as a multiclass classification problem. 
This approach invariably leads to a $\sqrt{pd}$ dependence on the number of parameters $pd$, with an additional factor of $\sqrt{d}$ in the general convex case or a square root logarithmic dependence on the number of extreme points in the polyhedral case.
This is not surprising, as the Natarajan dimension measures the maximum cardinality set (from the collection of extreme points) that can be shattered by a linear hypothesis class with $pd$ parameters.

However, given some constraints on the linear hypothesis class (i.e., ridge or lasso regularization) one would hope to improve the dependence on the number of parameters leveraging relevant results on the regular Rademacher complexity of the constrained version of $\calH_{\mathrm{lin}}$. One would also hope that the dependency on the number of extreme points is also not relevant to the generalization bound. 
Interestingly, both the dependence on the number of parameters and the dependence on the number of extreme points may be improved if the SPO loss were $L$-Lipschitz (w.r.t. the $\ell_2$-norm) due to the vector concentration inequality of \cite{maurer2016vector}, which would imply that $ \RadHatSPO(\calH)  $ is at most $ \sqrt{2} L \RadHat(\calH)$. Although the Rademacher complexity of a multivariate hypothesis class is less commonly used in machine learning, we show in Section \ref{sec:ambuj_bounds} that many of the results that bring down the dependency on the number of parameters in the univariate case still apply in the multivariate case.

Unfortunately, the  SPO loss is $\emph{not}$ Lipschitz. It is not so difficult to see that there can be sets $S$ with predictions $\hat{c}$ that yield wildly different solutions if $\hat{c}$ is perturbed, i.e., $w^\ast(\hat{c})$ is not Lipschitz. Furthermore, the SPO loss is actually discontinuous when $\hat{c}$ is a vector that has multiple optimal solutions (e.g., when $\hat{c}$ is the zero vector or a normal vector to a face of a polyhedron). On the other hand, the SPO loss may have desirable continuity properties when $\hat{c}$ is bounded away from the ``degenerate" cost vectors that lead to multiple optimal solutions.
This intuition motivates us to use a modified version of the SPO loss called the\textit{ margin SPO loss}. Our developments are akin to and in fact are a strict generalization of the margin analysis  for  classification  developed in \cite{koltchinskii2002empirical}. In particular, we extend the notion of margin from classification problems to the predict-then-optimize setting by associating the ``margin" of a cost vector prediction $\hat{c}$ to its ``distance to degeneracy," where here degeneracy refers to when problem \eqref{poi} has multiple optimal solutions.  The margin SPO loss imposes an additional penalty on cost vector predictions when they are close to being degenerate. We prove that the margin SPO loss is a Lipschitz function when $S$ satisfies what we call the\textit{ strength property}, which measures the sensitivity of the optimal solution around $\hat{c}$ with respect to the distance to degeneracy. We leverage the Lipschitz property of the margin SPO loss in order to develop improved generalization bounds that use the multivariate Rademacher complexity of $\calH$ as discussed above, at the small expense of using  the empirical margin SPO loss rather than the empirical SPO loss as in Theorem \ref{SPO_rademacher}. 

The analysis and results in this section hold generally for any convex and compact feasible region $S$ that satisfies the strength property, but the ability to compute the distance to degeneracy and subsequently evaluate the empirical margin SPO loss depends on the structural properties of $S$. We note also that the Lipschitz parameter of the margin SPO loss depends on this strength property and the size of the margin. A smaller margin implies a larger Lipschitz parameter yet a closer imitation of the margin SPO loss to the original SPO loss. In Section \ref{sec:margin-special}, we discuss two special cases, strongly convex sets and polyhedral sets with known convex hull representations, where the strength property is satisfied and where one can efficiently compute the distance to degeneracy, enabling the application of the generalization bounds developed herein.

\subsection{Margin SPO Loss}
Let us now present the definition of the set of degenerate cost vectors and the distance to degeneracy. Recall that $\|\cdot\|$ is a generic given norm on $\bbR^d$ with dual norm denoted by $\|\cdot\|_\ast$.
\begin{definition}[Distance to Degeneracy]\label{def:degeneracy}
The set of \emph{degenerate} cost vector predictions is $\calC^{\circ} := \{\hat{c} \in \bbR^d : P(\hat{c})$ has multiple optimal solutions$\}$, and the distance function (measured in the dual norm) to this set is denoted by $\nu_S(\hat{c}) := \dist_{\calC^{\circ}}^\ast(\hat{c}) = \inf_{c \in \calC^{\circ}}\left\{\|c - \hat{c}\|_\ast\right\}$. \Halmos
\end{definition}

\begin{figure}
\FIGURE{\centering
	\includegraphics[width=0.99\linewidth]{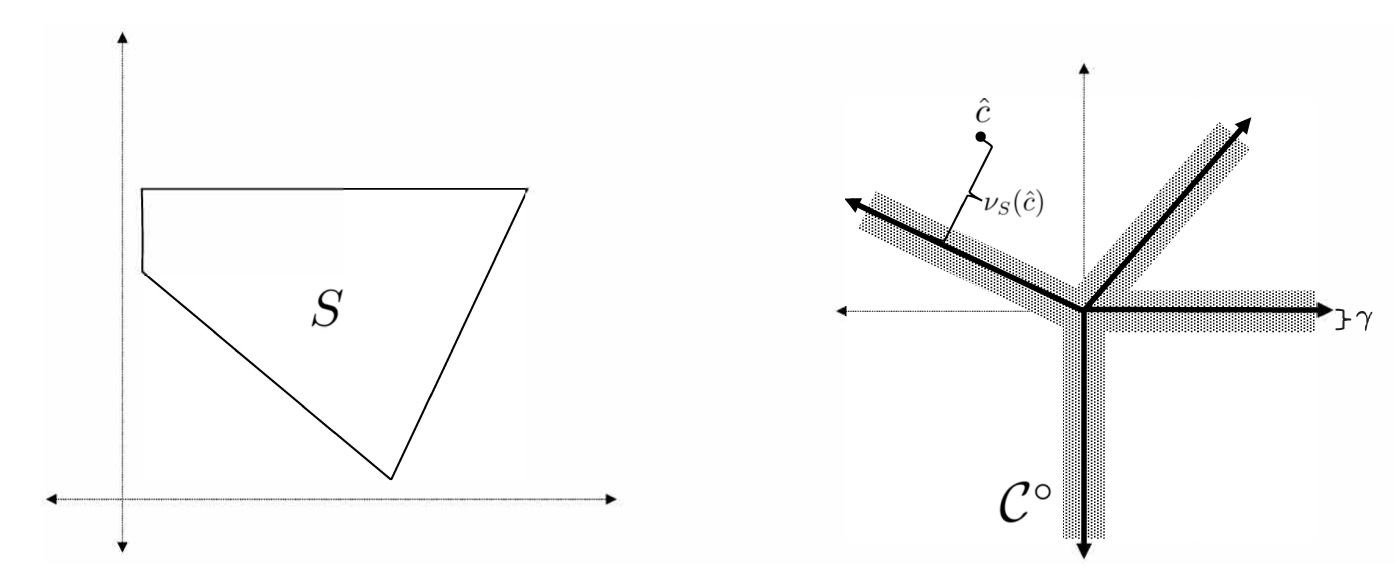}\label{fig}
}{\edit{A geometric example of distance to degeneracy.}}
{\edit{The left figure shows an example of a two-dimensional feasible region $S$ that corresponds to a polyhedron. The right figure shows the set of degenerate cost vectors $\calC^{\circ}$ in the black arrows. The shaded area around $\mathcal{C}^0$ represents a $\gamma$-margin around the degenerate set of cost vectors. For some cost vector $\hat{c}$, we show the distance to the closest degenerate point which is exactly $\nu_S(\hat{c})$ when using the 2-norm (whose dual norm is also the 2-norm).}} 
\end{figure}

Notice that the set $\calC^{\circ}$ and hence the function $\nu_S(\cdot)$ is determined completely by the set $S$, \edit{see Figure \ref{fig} for an example}. In this section, we assume without loss of generality that $S$ is not a singleton. Indeed, if $S$ is a singleton then the SPO loss $\ellT(\cdot, c)$ is identically equal to 0 for all $c \in \bbR^d$ and generalization bounds are trivial. Since $S$ is not a singleton, we have that $0 \in \calC^{\circ}$ and therefore $\nu_S(\hat{c}) \in [0, \infty)$ for all $\hat{c} \in \bbR^d$. (In fact, $0 \in \calC^{\circ}$ implies that $\nu_S(\hat{c}) \leq \|\hat{c}\|_\ast$.)
Therefore, since $\nu_S(\cdot)$ is a distance function to a non-empty set, it is also a 1-Lipschitz function (see, e.g., \cite{rockafellar2009variational} for a reference on this standard fact).
A key property of the feasible region $S$ and the function $\nu_S(\cdot)$ that will enable us to prove a ``Lipschitz-like" property of the optimization oracle $w^\ast(\cdot)$ is the following strength property formalized in Definition \ref{assumption:strength} below.
\begin{definition}[Strength Property]\label{assumption:strength}
The feasible region $S$ satisfies the \textit{strength property} if for some constant $\mu > 0$,
\begin{equation}\label{strength_property}
\hat{c}^T(w - w^\ast(\hat{c})) ~\geq~ \left(\frac{\mu\cdot\nu_S(\hat{c})}{2}\right)\|w - w^\ast(\hat{c})\|^2 \ \text{ for all } w \in S \text{ and } \hat{c} \in \bbR^d \ ,
\end{equation}
 where $\nu_S(\cdot)$ is the distance to degeneracy function. We refer to $\mu$ as the \textit{strength parameter}. \Halmos
\end{definition}


The strength property above may be thought of as a type of ``quantitative optimality condition" that justifies the use of the term ``strength." Indeed, \eqref{strength_property} indicates that larger values of $\nu_S(\hat{c})$ provide stronger bounds on the optimality gap $\hat{c}^T(w - w^\ast(\hat{c}))$ relative to the squared distance $\|w - w^\ast(\hat{c})\|^2$. \edit{The distance to degeneracy function is chosen to make the strength property as tight as possible, and thus ensuring better generalization bounds since we want to avoid points in the $\gamma$-margin of $\nu_S(\hat{c})$  as much as possible.} The name is also inspired by results in the case when $S$ is strongly convex, although in Section \ref{sec:margin-special} it it shown that the strength property holds in the polyhedral case as well. An example of a convex set that does not satisfy the strength property for any $\mu > 0$ is an $\ell_p$ ball where $2 < p < \infty$.

\begin{remark}[Normal Cone Interpretation]
Notice that the strength property in Definition \ref{assumption:strength} is independent of the optimization oracle $w^\ast(\cdot)$ since, for any $\hat{c}$ where there are multiple choices for $w^\ast(\hat{c})$, by definition it holds that $\nu_S(\hat{c}) = 0$ and thus \eqref{strength_property} is satisfied. In other words, we can equivalently restate \eqref{strength_property} as follows.
Let the normal cone of $S$ at a point $\bar{w} \in S$ be denoted by $N_S(\bar{w}) := \{c \in \bbR^d : c^T(w - \bar{w}) \leq 0 \text{ for all } w \in S\}$. Then, \eqref{strength_property} is equivalent to requiring that 
\begin{equation*}
N_S(\bar{w}) = \left\{c \in \bbR^d : c^T(w - \bar{w}) \leq -\left(\tfrac{\mu\cdot\nu_S(-c)}{2}\right)\|w - \bar{w}\|^2 \text{ for all } w \in S\right\} \ \text{ for all } \bar{w} \in S \ .
\end{equation*}
In Section \ref{sec:margin-special}, we show that strongly convex and polyhedral sets satisfy the strength property and thus the equivalent property above.
The above is a stronger characterization of the normal cones of $S$ and may be of interest in other contexts as well. For example, for \edit{the differentiable nonlinear minimization problem $\min_{w \in S} F(w)$}, the above gives a tighter characterization of the optimality condition $-\nabla F(\bar{w}) \in N_S(\bar{w})$. \Halmos
\end{remark}


As discussed, the distance to degeneracy $\nu_S(\hat{c})$ provides a measure of ``confidence'' regarding the cost vector prediction $\hat{c}$ and its implied decision $w^\ast(\hat{c})$. This intuition motivates us to define a margin-based version of the SPO loss, which places a greater penalty on cost vector predictions that are nearly degenerate. In other words, the ``margin SPO loss" encourages predictions that have small SPO loss values as well as predictions that are far enough away from the set of degenerate cost vector predictions $\calC^{\circ}$.
Definition \ref{def:margin_spo} presents the definition of the ``$\gamma$-margin SPO loss,'' which requires a fixed parameter $\gamma > 0$ that controls the size of the ``margin" around  $\calC^{\circ}$ where cost vector predictions are more heavily penalized.
\begin{definition}[$\gamma$-margin SPO Loss]\label{def:margin_spo}
For a fixed parameter $\gamma > 0$, given a cost vector prediction $\hat{c}$ and a realized cost vector $c$, the \emph{$\gamma$-margin SPO loss} $\SPOmar{\hat{c}}$ is defined as:
\begin{equation*}
\SPOmar{\hat{c}} :=
\begin{cases}
\ellT(\hat{c}, c) &\text{ if } \nu_S(\hat{c}) > \gamma \\
\left(\frac{\nu_S(\hat{c})}{\gamma}\right)\ellT(\hat{c}, c) + \left(1 - \frac{\nu_S(\hat{c})}{\gamma}\right)\ldiam(c) &\text{ if } \nu_S(\hat{c}) \leq \gamma. 
\Halmos
\end{cases}
\end{equation*} 
\end{definition}
Recall that, for any $\hat{c}, c \in \bbR^d$, it holds that $\ellT(\hat{c}, c) \leq \ldiam(c)$. Hence, we also have that $\ellT(\hat{c}, c) \leq \SPOmar{\hat{c}}$, that is the $\gamma$-margin SPO loss provides an upper bound on the SPO loss. Notice that the $\gamma$-margin SPO loss linearly interpolates between the SPO loss $\ellT(\hat{c}, c)$ and the upper bound $\ldiam(c)$ whenever $\nu_S(\hat{c}) \leq \gamma$. The $\gamma$-margin SPO loss also satisfies a simple monotonicity property whereby $\ellTmar(\hat{c}, c) \leq \ellT^{\bar{\gamma}}(\hat{c}, c)$ for any $\hat{c}, c \in \bbR^d$ and $\bar{\gamma} \geq \gamma > 0$.

We are now ready to state a theorem concerning the Lipschitz properties of the optimization oracle $w^\ast(\cdot)$ and the $\gamma$-margin SPO loss, which will then be used to derive margin-based generalization bounds. Theorem \ref{thm:main_lipschitz} below first demonstrates that the optimization oracle $w^\ast(\cdot)$ and SPO loss both satisfy a ``Lipschitz-like'' property when the distance to degeneracy function $\nu_S(\cdot)$ is bounded away from zero. Subsequently, these Lipschitz-like properties are a key ingredient in demonstrating that the $\gamma$-margin SPO loss is a Lipschitz function.

\begin{theorem}\label{thm:main_lipschitz}
Suppose that the feasible region $S$ satisfies the strength property with parameter $\mu$. Then, we have the following:
\begin{enumerate}
\item[(a)] The optimization oracle $w^\ast(\cdot)$ satisfies the following Lipschitz-like property: 
\begin{equation*}
\|w^\ast(\hat{c}_1) - w^\ast(\hat{c}_2)\| ~\leq~ \left(\frac{1}{\mu\cdot\min\{\nu_S(\hat{c}_1), \nu_S(\hat{c}_2)\}}\right)\|\hat{c}_1 - \hat{c}_2\|_\ast \ \text{ for all } \hat{c}_1, \hat{c}_2 \in \bbR^d.
\end{equation*}
\item[(b)] For any fixed $c \in \bbR^d$, the SPO loss satisfies the following Lipschitz-like property:
\begin{equation*}
|\ellT(\hat{c}_1, c) - \ellT(\hat{c}_2, c)| ~\leq~ \left(\frac{\|c\|_\ast}{\mu\cdot\min\{\nu_S(\hat{c}_1), \nu_S(\hat{c}_2)\}}\right)\|\hat{c}_1 - \hat{c}_2\|_\ast \ \text{ for all } \hat{c}_1, \hat{c}_2 \in \bbR^d.
\end{equation*}
\item[(c)] For any fixed $c \in \bbR^d$ and $\gamma > 0$, the $\gamma$-margin SPO loss is $\frac{1}{\gamma\mu}\left(\|c\|_\ast +  \mu\cdot\ldiam(c)\right)$-Lipschitz with respect to the dual norm $\|\cdot\|_\ast$, i.e., it holds that:
\begin{equation*}
|\SPOmar{\hat{c}_1} - \SPOmar{\hat{c}_2}| ~\leq~ \left(\frac{\|c\|_\ast +  \mu\cdot\ldiam(c)}{\gamma\mu}\right)\|\hat{c}_1 - \hat{c}_2\|_\ast \ \text{ for all } \hat{c}_1, \hat{c}_2 \in \bbR^d \ .
\end{equation*}
\end{enumerate}
\end{theorem}
\proof{Proof.} If $S$ is a singleton, then all of these results are trivial. Otherwise, as mentioned previously, $\nu_S(\cdot)$ is finite valued and 1-Lipschitz on $\bbR^d$. Let $\hat{c}_1, \hat{c}_2 \in \bbR^d$ be given and let $\tau := \min\{\nu_S(\hat{c}_1), \nu_S(\hat{c}_2)\}$. 
\paragraph{Part (a):}
If $\tau = 0$ then (a) holds since the right-hand side of (a) is equal to $+\infty$ by convention. Otherwise $\tau > 0$ and applying  \eqref{strength_property} twice yields
\begin{equation*}
\hat{c}_1^T(w^\ast(\hat{c}_2) - w^\ast(\hat{c}_1)) ~\geq~ \left(\tfrac{\mu\cdot\nu_S(\hat{c}_1)}{2}\right)\|w^\ast(\hat{c}_2) - w^\ast(\hat{c}_1)\|^2 ~\geq~ \left(\tfrac{\mu\tau}{2}\right)\|w^\ast(\hat{c}_1) - w^\ast(\hat{c}_2)\|^2 \ ,
\end{equation*}
and
\begin{equation*}
\hat{c}_2^T(w^\ast(\hat{c}_1) - w^\ast(\hat{c}_2)) ~\geq~ \left(\tfrac{\mu\cdot\nu_S(\hat{c}_2)}{2}\right)\|w^\ast(\hat{c}_1) - w^\ast(\hat{c}_2)\|^2 ~\geq~ \left(\tfrac{\mu\tau}{2}\right)\|w^\ast(\hat{c}_1) - w^\ast(\hat{c}_2)\|^2 \ .
\end{equation*}
Adding the above two inequalities together yields
\begin{equation*}
\mu\tau\|w^\ast(\hat{c}_1) - w^\ast(\hat{c}_2)\|^2 ~\leq~ (\hat{c}_2 - \hat{c}_1)^T(w^\ast(\hat{c}_1) - w^\ast(\hat{c}_2)) ~\leq~ \|\hat{c}_1 - \hat{c}_2\|_\ast\|w^\ast(\hat{c}_1) - w^\ast(\hat{c}_2)\| \ ,
\end{equation*}
where the second inequality is H{\"o}lder's inequality. Dividing both sides of the above by $\mu\tau\|w^\ast(\hat{c}_1) - w^\ast(\hat{c}_2)\|$ yields item (a). 

\paragraph{Part (b):} Now let $c \in \bbR^d$ be given and note that
\begin{equation*}
|\ellT(\hat{c}_1, c) - \ellT(\hat{c}_2, c)| ~=~ |c^T(w^\ast(\hat{c}_1) - w^\ast(\hat{c}_2))| ~\leq~ \|c\|_\ast\|w^\ast(\hat{c}_1) - w^\ast(\hat{c}_2)\| \ ,
\end{equation*}
where the inequality follows from  H{\"o}lder's inequality. Hence, (b) follows by combining the above with (a).

\paragraph{Part (c):}
 Without loss of generality, we consider three cases:  {\em (i)} $\nu_S(\hat{c}_1) \leq \gamma$ and $\nu_S(\hat{c}_2) \leq \gamma$, {\em (ii)} $\nu_S(\hat{c}_1) \geq \gamma$ and $\nu_S(\hat{c}_2) \geq \gamma$, and {\em (iii)} $\nu_S(\hat{c}_1) \leq \gamma$ and $\nu_S(\hat{c}_2) > \gamma$. \

Let us first consider case {\em (i)} and let us  suppose without loss of generality that $0 < \nu_S(\hat{c}_1) \leq \nu_S(\hat{c}_2) \leq \gamma$. By definition, we know that $|\SPOmar{\hat{c}_1} - \SPOmar{\hat{c}_2}|$ is equal to
\begin{align*}
&\left|  \left(\frac{\nu_S(\hat{c}_1)}{\gamma}\right)\ellT(\hat{c}_1, c) + \left(1 - \frac{\nu_S(\hat{c}_1)}{\gamma}\right)\ldiam(c)  -  \left(\frac{\nu_S(\hat{c}_2)}{\gamma}\right)\ellT(\hat{c}_2, c)-  \left(1 - \frac{\nu_S(\hat{c}_2)}{\gamma}\right)\ldiam(c) \right| \\
=& \left|\left(\frac{\nu_S(\hat{c}_1)}{\gamma}\right)[\ellT(\hat{c}_1, c) - \ellT(\hat{c}_2, c)] + \left(\frac{\ellT(\hat{c}_2, c) - \ldiam(c)}{\gamma}\right)[\nu_S(\hat{c}_1) - \nu_S(\hat{c}_2)]    \right| \\
\leq & \left(\frac{\nu_S(\hat{c}_1)}{\gamma}\right) \left| \ellT(\hat{c}_1, c) - \ellT(\hat{c}_2, c)  \right| +  \left(\frac{\ldiam(c)}{\gamma}\right) \left|  \nu_S(\hat{c}_1) - \nu_S(\hat{c}_2)     \right| \\
 \leq & \left(\frac{\nu_S(\hat{c}_1)}{\gamma}\right)\left(\frac{\|c\|_\ast}{\mu\cdot\nu_S(\hat{c}_1)}\right)\|\hat{c}_1 - \hat{c}_2\|_\ast + \left(\frac{\ldiam(c)}{\gamma}\right)\|\hat{c}_1 - \hat{c}_2\|_\ast \\
= &  \frac{1}{\gamma}\left(\frac{\|c\|_\ast}{\mu} + \ldiam(c)\right)\|\hat{c}_1 - \hat{c}_2\|_\ast \ ,
\end{align*}
where the first inequality follows from the triangle inequality and the definition of $\ldiam(c)$ and the second inequality follows from part (b) and the facts that $0 < \nu_S(\hat{c}_1) = \min\{\nu_S(\hat{c}_1), \nu_S(\hat{c}_2)\}$ and $\nu_S(\cdot)$ is 1-Lipschitz.



Now, in case {\em (ii)}, we have that $\SPOmar{\hat{c}_1} = \ellT(\hat{c}_1, c)$ and $\SPOmar{\hat{c}_2} = \ellT(\hat{c}_2, c)$.
Therefore, applying  part (b) yields
\begin{equation*}
|\SPOmar{\hat{c}_1} - \SPOmar{\hat{c}_2}| \leq   \frac{\|c\|_\ast}{\mu\cdot\min\{\nu_S(\hat{c}_1), \nu_S(\hat{c}_2)\}} \|\hat{c}_1 - \hat{c}_2\|_\ast \leq  \frac{\|c\|_\ast}{\gamma\mu}\|\hat{c}_1 - \hat{c}_2\|_\ast \leq    \frac{\|c\|_\ast +  \mu\cdot\ldiam(c)}{\gamma\mu}  \|\hat{c}_1 - \hat{c}_2\|_\ast  \ ,
\end{equation*}
where the second  inequality follows since $\gamma \leq \min\{\nu_S(\hat{c}_1), \nu_S(\hat{c}_2)\}$.

Finally, in case {\em (iii)}, by the intermediate value theorem we can define $\bar{c} := \lambda\hat{c}_1 + (1 - \lambda)\hat{c}_2$ where $\lambda \in (0,1]$   such that $\nu_S(\bar{c}) = \gamma$. Then, we have that $  |\SPOmar{\hat{c}_1} - \SPOmar{\hat{c}_2}|$ is equal to
\begin{align*}
 &|(\SPOmar{\hat{c}_1} - \SPOmar{\bar{c}}) + (\SPOmar{\bar{c}} - \SPOmar{\hat{c}_2})| \\
~ \leq &~ |\SPOmar{\hat{c}_1} - \SPOmar{\bar{c}}| + |\SPOmar{\bar{c}} - \SPOmar{\hat{c}_2}| \\
~ \leq &~ \left(\frac{\|c\|_\ast +  \mu\cdot\ldiam(c)}{\gamma\mu}\right)\|\hat{c}_1 - \bar{c}\| + \left(\frac{\|c\|_\ast +  \mu\cdot\ldiam(c)}{\gamma\mu}\right)\|\bar{c} - \hat{c}_2\| \\
~ = &~ \left(\frac{\|c\|_\ast +  \mu\cdot\ldiam(c)}{\gamma\mu}\right)(\|\hat{c}_1 - \bar{c}\| + \|\bar{c} - \hat{c}_2\|)  ,
\end{align*}
where the first inequality follows from the triangle inequality, the second inequality follows from cases {\em (i)} and {\em (ii)}, and the final equality follows the definition of $\bar{c}$  \Halmos \endproof 

\subsection{Connecting to Multivariate Rademacher Complexity} 
As we are extending standard margin-based generalization theory from binary classification to the predict-then-optimize setting, it is instructive to recall the steps involved in the binary classification setting (e.g., see \cite[Section 5.4]{mohri2018foundations}). First, the $0$-$1$ loss function is upper bounded by the $\gamma$-margin loss function. Second, the Rademacher complexity of the $\gamma$-margin loss function class is upper bounded by the Rademacher complexity of the underlying function class. This second step uses the Lipschitz property of the $\gamma$-margin loss function and a result known as the Ledoux-Talagrand contraction inequality (e.g., see \cite[Lemma 5.7]{mohri2018foundations})

We have already accomplished step one above by constructing the $\gamma$-margin SPO loss which is Lipschitz and also upper bounds the SPO loss. Note that this step was significantly more challenging than in the binary classification case.
Step two also poses its own set of challenges. In the binary classification case, the function class is a class of scalar-valued functions. However, in our case, the class $\calH$ consists of {\em vector-valued functions}. This raises two important questions. First, how do we define the Rademacher complexity of a vector-valued function class?
Second, what is the analogue of the Ledoux-Talagrand contraction inequality for the vector case?

Recall that the multivariate empirical Rademacher complexity of $\calH$ is defined as
\begin{equation}\label{rad_multi}
\RadHat(\calH) := \bbE_{\sigma}\left[\sup_{f \in \calH}\frac{1}{n}\sum_{i = 1}^n\sum_{j = 1}^d \sigma_{ij}f_j(x_i)\right] =
\bbE_{\vsigma}\left[\sup_{f \in \calH}\frac{1}{n}\sum_{i = 1}^n \vsigma_{i}^T f(x_i)\right] \ ,
\end{equation}
where $\sigma_{ij}$ are i.i.d.\ Rademacher random variables for $i = 1, \ldots, n$ and $j = 1, \ldots, d$, and $\vsigma_i := (\sigma_{i1},\ldots,\sigma_{id})^T$. The expected version of the multivariate Rademacher complexity is defined as $\Rad(\calH) := \bbE\left[\RadHat(\calH)\right]$ where the expectation is taken with respect to the i.i.d.\ sample drawn from the underlying distribution $\calD$.
It is often the case that the structure of the hypothesis class $\calH$ naturally leads to a bound on $\Rad(\calH)$ that can have mild, even logarithmic, dependence on dimensions $p$ and $d$. We give examples of such bounds in Section \ref{sec:ambuj_bounds}, after first presenting our margin-based generalization bounds.

Let us also introduce the empirical $\gamma$-margin SPO loss:
\begin{equation*}
\empRiskTmar(f) := \frac{1}{n}\sum_{i = 1}^n\ellTmar(f(x_i), c_i) \ ,
\end{equation*}
and the empirical Rademacher complexity of $\calH$ with respect to the $\gamma$-margin SPO loss:
\begin{equation*}
  \RadHatSPOmar(\calH) := \bbE_{\sigma}\left[\sup_{f \in \calH}\frac{1}{n}\sum_{i = 1}^n \sigma_{i}\ellTmar(f(x_i), c_i)\right] \ ,
\end{equation*}
where $f \in \calH$ on the left side above and $\sigma_{i}$ are i.i.d.\ (scalar) Rademacher random variables for $i = 1, \ldots, n$. So we need tools that allow us to transition from the standard Rademacher complexity of $\calH$ with respect to the $\gamma$-margin SPO loss to the multivariate Rademacher complexity of $\calH$. 

Let us now briefly review the concept of vector contraction inequalities for multivariate Rademacher complexities, which turns out to provide precisely the right set of tools. Let $\Phi_i:\bbR^d\to\bbR$ for $i \in \{1, \ldots, n\}$ be a collection of $L$-Lipschitz functions with respect to the given norm $\| \cdot \|$ defined on $\bbR^d$:
\[
| \Phi_i(u) - \Phi_i(v) | 
\le L \cdot \| u - v \| \ \text{ for all } u, v \in \bbR^d.
\] 
A \emph{vector contraction inequality} takes the form:
\begin{equation}\label{general_vec_contract}
    \bbE_{\sigma}\left[\sup_{f \in \calH}\frac{1}{n}\sum_{i = 1}^n \sigma_{i} \Phi_i(f(x_i))\right]
    \le
    CL \cdot
    \bbE_{\vsigma}\left[\sup_{f \in \calH}\frac{1}{n}\sum_{i = 1}^n \vsigma_{i}^T f(x_i) \right]
    =
    CL\cdot\RadHat(\calH) \ ,
\end{equation}
where $C$ is a constant. As we have already demonstrated that the $\gamma$-margin SPO loss is Lipschitz, our next step is to apply a vector contraction inequality in order to obtain improved generalization bounds.

\subsection{Generalization Bounds in the $\ell_2$ Case}
We now focus on the case of the $\ell_2$-norm set-up, i.e., the norm on the space of $w$ variables as well as the norm on the space of cost vectors $c$ are both the $\ell_2$-norm. Our primary reason for focusing on the $\ell_2$ case is that we can apply an elegant vector contraction inequality due to \cite{maurer2016vector}, which exactly takes the form of \eqref{general_vec_contract} with $C = \sqrt{2}$. Vector contraction inequalities have also been developed in other norm setups, but the result of \cite{maurer2016vector} in the $\ell_2$ case appears to be the most elegant and the most applicable result for our purposes. 

\begin{remark}[Other Vector Contraction Inequalities] Indeed, for the case when $\Phi_i$ are Lipschitz with respect to the $\ell_\infty$-norm, the first relevant inequality is that of \cite{bertsimas2014predictive} (see Lemma EC.1 in their supplementary material). Note that this inequality is inferior to the result of \cite{maurer2016vector} since the $\ell_\infty$ Lipschitz constant of a function is always larger than, or equal to, the $\ell_2$ Lipschitz constant. An alternative contraction inequality for the $\ell_\infty$ case is due to \citet{foster2019vector}, which involves a worst case alternative to the multivariate Rademacher complexity and may be favorable in some situations. Finally, \citet{zatarain2019vector} has recently considered the case of an $\ell_p$-norm for $p \in (1,2)$, which involves replacing Rademacher random variables with $p$-stable random variables. The result of \citet{zatarain2019vector} is not particularly useful in the case of strongly convex sets $S$ since all common examples of strongly convex sets with strong convexity constant independent of dimension are strongly convex with respect to an $\ell_p$ norm for some $p \le 2$. However, this result may have potential uses for some polyhedral sets. \Halmos
\end{remark}

With Theorem \ref{thm:main_lipschitz} and the vector contraction inequality of \cite{maurer2016vector} in place, we are now ready to present our margin-based generalization bound in the predict-then-optimize setting in Theorem \ref{thm:margin_generalization} below.
Recall that $\calC$ denotes the domain of the true cost vectors $c$, $\rho_2(\calC) := \sup_{c \in \calC}\|c\|_2$, and $\ldiam(\calC) := \sup_{c \in \calC}\ldiam(c)$.

\begin{theorem}\label{thm:margin_generalization}
Suppose that the feasible region $S$ satisfies the strength property with parameter $\mu > 0$ and with respect to the $\ell_2$-norm. Let $\gamma > 0$ be fixed, and let $\calH$ be a family of functions mapping from $\calX$ to $\bbR^d$. Then, for any fixed sample $ ((x_1,c_1)...(x_n,c_n))$ we have that
\begin{equation*}
\RadHatSPOmar(\calH) ~\leq~
\left(\frac{\sqrt{2}\rho_2(\calC) + \sqrt{2}\mu\cdot\ldiam(\calC)}{\gamma\mu}\right)\RadHat(\calH) \ .
\end{equation*}
Furthermore, for any $\delta > 0$, with probability at least $1 - \delta$ over an i.i.d. sample drawn from the distribution $\calD$, the following holds for all $f \in \calH$
\begin{equation*}
\riskT(f) ~\leq~ \empRiskTmar(f) + \left(\frac{2\sqrt{2}\rho_2(\calC) + 2\sqrt{2}\mu\cdot\ldiam(\calC)}{\gamma\mu}\right)\Rad(\calH) + \ldiam(\calC)\sqrt{\frac{\log(1/\delta)}{2n}} \ .
\end{equation*}
\end{theorem}
\proof{Proof.}
The bound on $\RadHatSPOmar(\calH)$ follows simply by combining   Theorem \ref{thm:main_lipschitz}(c) with the vector contraction inequality \eqref{general_vec_contract} due to \cite{maurer2016vector}. The subsequent generalization bound then simply follows since $\riskT(f) \leq \riskTmar(f)$ for all $f \in \calH$ and by applying the version of Theorem \ref{SPO_rademacher} for the $\gamma$-margin SPO loss.
\Halmos \endproof

Theorem \ref{thm:margin_generalization} may also be extended to a bound that holds uniformly over all values of $\gamma \in (0, \bar{\gamma}]$, where $\bar{\gamma} > 0$ is a fixed parameter. This extension is presented below in Theorem \ref{thm:margin_generalization_uniform} and proved in Appendix \ref{sec:thm5proof}. The advantage of the bound presented in Theorem \ref{thm:margin_generalization_uniform} versus the bound presented in Theorem \ref{thm:margin_generalization} is that $\gamma$ in Theorem \ref{thm:margin_generalization_uniform} can be chosen in a data-driven way, e.g., through cross-validation.

\begin{theorem}\label{thm:margin_generalization_uniform}
Suppose that the feasible region $S$ satisfies the strength property with parameter $\mu > 0$ and with respect to the $\ell_2$-norm. Let $\bar\gamma > 0$ be fixed, and let $\calH$ be a family of functions mapping from $\calX$ to $\bbR^d$.
Then, for any $\delta > 0$, with probability at least $1 - \delta$ over an i.i.d.\ sample drawn from the distribution $\calD$, the following holds for all $f \in \calH$ and for all $\gamma \in (0, \bar\gamma]$
\begin{equation*}
\riskT(f) \leq \empRiskTmar(f) + \left(\frac{4\sqrt{2}\rho_2(\calC) + 4\sqrt{2}\mu\cdot\ldiam(\calC)}{\gamma\mu}\right)\Rad(\calH) + \ldiam(\calC)\left(\sqrt{\frac{\log(\log_2(2\bar\gamma/\gamma))}{n}} + \sqrt{\frac{\log(2/\delta)}{2n}}\right) \ .
\end{equation*}
\end{theorem}
The proof of Theorem \ref{thm:margin_generalization_uniform} is presented in Appendix \ref{sec:uniform_proof}.
Note that a natural choice for $\bar\gamma$ in Theorem \ref{thm:margin_generalization_uniform} is $\bar\gamma \gets \sup_{f \in \calH, x \in \calX}\nu_S(f(x))$ or an upper bound thereof, presuming that one can obtain such a bound based on the properties of $S$, $\calH$, and $\calX$. In addition, since $\nu_S(\hat{c}) \leq \|\hat{c}\|_2$ for all $\hat{c} \in \bbR^d$, one can also take $\bar\gamma \gets \sup_{f \in \calH, x \in \calX}\|f(x)\|_2$ or an upper bound thereof. As mentioned, the value of $\gamma$ in Theorem \ref{thm:margin_generalization_uniform} can be chosen in a data-driven way so that, given a prediction function $\hat{f}$ trained on the observed data, the upper bound on $\empRiskT(\hat{f})$ given by Theorem \ref{thm:margin_generalization_uniform} is minimized. Since Theorem \ref{thm:margin_generalization_uniform} is a uniform result over $\gamma \in (0, \bar{\gamma}]$, this data-driven procedure for choosing $\gamma$ is indeed valid.

It is important to note that the utility of Theorems \ref{thm:margin_generalization} and \ref{thm:margin_generalization_uniform} is strengthened when the underlying distribution $\calD$ has a ``favorable margin property,'' whereby $\empRiskTmar(f) \approx \empRiskT(f)$ for some value of $\gamma > 0$ that is reasonably large. A sufficient condition to ensure such a favorable margin property is if $\nu_S(f(x)) \geq \gamma$ with high probability over $x$, i.e., most predictions of the model $f$ are far from degenerate. When such a favorable margin property occurs, the bounds in Theorems \ref{thm:margin_generalization} and \ref{thm:margin_generalization_uniform} can be much stronger than those of Corollaries \ref{cor:linear} and \ref{cor:balls}. Indeed, we demonstrate in Section  \ref{sec:ambuj_bounds} and \ref{sec:margin-special} examples of hypothesis classes and feasible regions, respectively, such that the dependence on the number of parameters $pd$ in the bounds of Theorems \ref{thm:margin_generalization} and \ref{thm:margin_generalization_uniform} is \edit{significantly stronger than that of Corollaries \ref{cor:linear} and \ref{cor:balls}.}

\subsection{Bounding the Multivariate Rademacher Complexity for Linear Classes}\label{sec:ambuj_bounds}

As mentioned, it is often the case that the structure of the hypothesis class $\calH$ naturally leads to a bound on $\Rad(\calH)$ that \edit{has favorable dependence} on dimensions $p$ and $d$. For example, let us consider the general setting of a constrained linear function class, namely $\calH = \calH_\calB := \{f \::\: f(x) = Bx \text{ for some } B \in \bbR^{d \times p}, B \in \calB\}$, where $\calB \subseteq \bbR^{d \times p}$. Theorems \ref{thm:margin_generalization} and \ref{thm:margin_generalization_uniform} apply more broadly than just linear function classes, but these examples are illustrative and comparable to Corollaries \ref{cor:linear} and \ref{cor:balls}.
Theorem \ref{thm:linear_class} below is an extension of Theorem 3 of \cite{kakade2009complexity} to multivariate Rademacher complexity and provides a convenient way to bound $\Rad(\calH_\calB)$ in the case when $\calB$ lies in a level set of a strongly convex function. In a slight abuse of notation, in Theorem \ref{thm:linear_class} only we use $\|\cdot\|$ and $\|\cdot\|_\ast$ to refer to a norm and corresponding dual norm, respectively, on the space of \emph{matrices} $\bbR^{d \times p}$. The full proof is in Appendix \ref{sec:thm6proof}.



\begin{theorem}\label{thm:linear_class}
Let $\calS \subseteq \bbR^{d \times p}$ be a closed convex set, and let $F: \calS \to \bbR$ be $\alpha$-strongly convex with respect to $\|\cdot\|_*$ and satisfy $\inf_{B \in \calS} F(B) = 0$. Suppose that $\calB \subseteq \{ B \in \calS \::\: F(B) \le \beta^2 \}$ and let $\Omega$ be such that
\[
\sup_{\vsigma \in \{\pm1\}^d} \sup_{x \in \calX} \| \vsigma x^T \| ~\le~ \Omega .
\]
Then, it holds that
\[
\Rad(\calH_\calB) ~\le~ \Omega \beta \sqrt{\frac{2}{\alpha n}} .
\]
\end{theorem}

This theorem can be applied with many different strongly convex functions of matrices \cite[Section 2.4]{kakade2012regularization}. We give some interesting examples below.

\begin{example}[Bounded Frobenius norm, i.e., ridge regularization]\label{ex:fro}
The most basic case is $\calB = \{ B \in \bbR^{d \times p} \::\: \tfrac{1}{2}\|B\|_F^2 \le \beta^2 \}$, in which case $F(B) = \tfrac{1}{2} \| B \|_F^2$ is $1$-strongly convex on $\bbR^{d \times p}$ w.r.t. $\| \cdot \|_F$. Note that 
\[
\sup_{\vsigma \in \{\pm1\}^d} \sup_{x \in \calX} \| \vsigma x^T \|_F =
\sup_{\vsigma \in \{\pm1\}^d} \| \vsigma \|_2 \cdot \sup_{x \in \calX} \| x \|_2
= \sqrt{d} \sup_{x \in \calX} \| x \|_2 .
\]
Therefore, if $\tfrac{1}{2}\| B \|_F^2 \le \beta^2$ and $\rho_2(\calX) := \sup_{x \in \cal X} \| x \|_2$ we have
\[
\Rad(\calH_\calB) ~\le~  \rho_2(\calX) \beta \sqrt{\frac{2d}{n}} .
\]
\Halmos \end{example}

\begin{example}[Bounded $\ell_1$ norm of vectorized matrix, i.e., lasso regularization]\label{ex:ell1}
Another case is when $\calB = \{ B \in \bbR^{d \times p} \::\: \tfrac{1}{2}\|B\|_1^2 \le \beta^2 \}$, where $\| B \|_q$ is $\ell_q$ norm of the vectorized matrix $B$.
We set $F(B) = \tfrac{1}{2} \| B \|_q^2$
for $q = \frac{\log(pd)}{\log(pd)-1}$ which is $1/(3\log(pd))$-strongly convex w.r.t. $\| \cdot \|_1$ \cite[Corollary 10]{kakade2012regularization}. Since $\| B \|_q \le \| B \|_1$, we clearly have $F(B) \le \beta^2$. 
Note that
\[
\sup_{\vsigma \in \{\pm1\}^d} \sup_{x \in \calX} \| \vsigma x^T \|_{\infty} = \sup_{\vsigma \in \{\pm1\}^d} \| \vsigma \|_\infty \cdot \sup_{x \in \calX} \| x \|_\infty
= \sup_{x \in \calX} \| x \|_\infty .
\]
Therefore, if $\tfrac{1}{2}\| B \|_1^2 \le \beta^2$ and $\rho_\infty(\calX) := \sup_{x \in \cal X} \| x \|_\infty$ we have
\[
\Rad(\calH_\calB) ~\le~ \rho_\infty(\calX) \beta \sqrt{\frac{6\log(pd)}{n}} .
\]
\Halmos \end{example}

\begin{example}[Bounded group-lasso norm]\label{ex:group}
In cases where the feature dimension $p$ is large, we might want to encode prior knowledge that only a subset of the $p$ input variables are relevant for making predictions. The vectorized $\ell_1$ norm considered in the previous example encourages sparsity but does not result in shared sparsity structure over the rows of $B$. That is, it does not cause entire columns to be set to zero. In multivariate regression, the group-lasso norm \cite[Section 4.3]{tibshirani2015statistical} is used to enforce such a structured from of sparsity. Define the norm
\[
\| B \|_{2,q} = \left( \sum_{j=1}^p \| B_{\cdot j} \|_2^q \right)^{1/q} .
\]
The subscripts above remind us that we first take the $\ell_2$ norms of columns $B_{\cdot j}$ and then take the $\ell_q$ norm of the $p$ resulting values. The group-lasso norm is simply $\| \cdot \|_{2,1}$. Let us consider the case when the matrices $B$ are constrained to have low group-lasso norm, i.e., $\calB = \{ B \in \bbR^{d \times p} \::\: \tfrac{1}{2} \| B \|_{2,1}^2 \le \beta^2 \}$.
We set $F(B) = \tfrac{1}{2} \| B \|_{2,q}^2$
for $q = \frac{\log(p)}{\log(p)-1}$ which is $1/(3\log(p))$-strongly convex w.r.t. $\| \cdot \|_{2,1}$ \cite[Corollary 14]{kakade2012regularization}. Since $\| B \|_{2,q} \le \| B \|_{2,1}$, we clearly have $F(B) \le \beta^2$. 
Note that the dual norm of the $\| \cdot \|_{2,1}$ norm is the $\| \cdot \|_{2,\infty}$ norm and we have
\begin{equation*}
\sup_{\vsigma \in \{\pm1\}^d} \sup_{x \in \calX} \| \vsigma x^T \|_{2,\infty}
=
\sup_{\vsigma \in \{\pm1\}^d} \| \vsigma \|_{2} \cdot \sup_{x \in \calX} \| x \|_\infty
\le \sqrt{d} \sup_{x \in \calX} \| x \|_\infty .
\end{equation*}
Therefore, if $\tfrac{1}{2}\| B \|_{2,1}^2 \le \beta^2$ and $\rho_\infty(\calX) := \sup_{x \in \cal X} \| x \|_\infty$ we have
\[
\Rad(\calH_\calB) ~\le~ \rho_\infty(\calX) \beta \sqrt{\frac{6 d \log(p)}{n}} .
\]
\Halmos \end{example}

\edit{
\begin{remark}[Dimension Dependence]
The reader will notice that in Examples~\ref{ex:fro} and~\ref{ex:group}, the dependence on the output/decision dimension $d$ is $O(\sqrt{d})$ in contrast to the dependence on input/feature dimension $p$ which is either absent (Example~\ref{ex:fro}) or logarithmic (Examples~\ref{ex:ell1} and~\ref{ex:group}). This is a direct consequence of the fact that multivariate Rademacher complexity involves a $d$-dimensional \emph{vector} of Radamacher variables \emph{per training example}. In standard Rademacher complexity, the output dimension is $1$ and therefore the dependence on $d$ is not manifest. However, it shows up in our more general setting.
\Halmos
\end{remark}
}

\section{Applications to Strongly Convex and Polyhedral Sets}\label{sec:margin-special}

In this section, we consider two special cases for the feasible region $S$:  {\em (i)} $S$ is a strongly convex set, and {\em (ii)} $S$ is polyhedral. In both cases, we show that the distance to degeneracy function $\nu_S(\cdot)$ satisfies the strength property and specify the strength parameter $\mu$. In the strongly convex case, $\nu_S(\hat{c}) = \|\hat{c}\|_\ast$ and is easily computable whenever the dual norm is. In the polyhedral case, we show that a sufficient condition for computing $\nu_S(\cdot)$ is that $S$ has a known convex hull representation, i.e., $S = \conv\{v_1, \ldots, v_K\}$ where $v_1, \ldots, v_K \in \bbR^d$ are given vectors.
Thus, in these two cases, we can readily compute the empirical margin SPO loss and apply Theorems \ref{thm:margin_generalization} and \ref{thm:margin_generalization_uniform}.

\subsection{Strongly Convex Sets}
We adopt the classical definition of strongly convex sets (see, e.g., \cite{vial1983strong, journee2010generalized, garber2015faster}), which is reviewed in Definition \ref{def:sc_set} below. Recall that $\|\cdot\|$ is a generic given norm on $\bbR^d$ with dual norm $\|\cdot\|_\ast$, and $B(\bar{w}, r) := \{w : \|w - \bar{w}\| \leq r\}$ denotes the ball of radius $r$ centered at $\bar{w}$. Recall also that the normal cone of $S$ at a point $\bar{w} \in S$ is defined by $N_S(\bar{w}) := \{c \in \bbR^d : c^T(w - \bar{w}) \leq 0 \text{ for all } w \in S\}$.
\begin{definition}[Strongly Convex Set]\label{def:sc_set}
For a constant $\bar{\mu} \geq 0$, we say that a convex set $S \subseteq \bbR^d$ is $\bar{\mu}$-strongly convex with respect to the norm $\|\cdot\|$ if, for any $w_1, w_2 \in S$ and for any $\lambda \in [0,1]$, it holds that:
\begin{equation*}
B\left(\lambda w_1 + (1 - \lambda)w_2, \left(\tfrac{\bar{\mu}}{2}\right)\lambda(1 - \lambda)\|w_1 - w_2\|^2\right) \subseteq S \ .  \Halmos
\end{equation*}
\end{definition}
Informally, Definition \ref{def:sc_set} says that, for every convex combination of points in $S$, a ball of appropriate radius also lies in $S$. Several examples of strongly convex sets are presented by \cite{journee2010generalized} and \cite{garber2015faster}, including $\ell_q$ and Schatten $\ell_q$ balls for $q \in (1,2]$, certain group norm balls, and generally any level set of a smooth and strongly convex function.

Our analysis herein relies on the following proposition, which gives a tighter characterization of the normal cones of strongly convex sets. Proposition \ref{prop:strong-normal} can be derived as a consequence of a result of \cite{vial1983strong}, but we include its proof in Appendix \ref{sec:prop1proof} in our notation for completeness.

\begin{proposition}[\cite{vial1983strong}, Proposition 2.9] \label{prop:strong-normal}
Let $S \subseteq \bbR^d$ be a $\bar{\mu}$-strongly convex set for some $\bar{\mu} \geq 0$. Then, for any $\bar{w} \in S$ it holds that:
\begin{equation}\label{eq:strong-normal}
N_S(\bar{w}) = \left\{c \in \bbR^d : c^T(w - \bar{w}) \leq -\left(\tfrac{\bar{\mu}}{2}\right)\|c\|_\ast\|w - \bar{w}\|^2 \text{ for all } w \in S\right\} \ .
\end{equation}
\end{proposition}

We are now ready to state our main theorem in the strongly convex case, which shows that the distance to degeneracy is exactly the dual norm and that \eqref{strength_property} is satisfied.
\begin{theorem} \label{thm:strongly}
Suppose that $S$ is is not a singleton and is $\bar{\mu}$-strongly convex for some $\bar{\mu} > 0$. Then the distance to degeneracy function satisfies $\nu_S(\hat{c}) = \|\hat{c}\|_\ast$ for all $\hat{c} \in \bbR^d$, and the strength property is satisfied with $\mu \gets \bar{\mu}$.
\end{theorem}
\proof{Proof.}
To show that $\nu_S(\hat{c}) = \|\hat{c}\|_\ast$, it suffices to show that $\calC^{\circ} = \{0\}$. Since $S$ is not a singleton, we have that $0 \in \calC^{\circ}$. Conversely, suppose that $\hat{c} \neq 0$. Then, since $\|\hat{c}\|_\ast > 0$ and $-\hat{c} \in N_S(w^\ast(\hat{c}))$, Proposition \ref{prop:strong-normal} implies that $\hat{c}^Tw^\ast(\hat{c}) < \hat{c}^Tw$ for all $w \in S$. Thus $w^\ast(\hat{c})$ is the unique optimal solution of $P(\hat{c})$ and $\hat{c} \not\in \calC^{\circ}$. Then the strength property with parameter $\bar{\mu}$ follows immediately from Proposition \ref{prop:strong-normal} and the observation that $-\hat{c} \in N_S(w^\ast(\hat{c}))$.
\Halmos \endproof

With Theorem \ref{thm:strongly} in place, we can now apply the margin-based generalization bounds of Theorems \ref{thm:margin_generalization} and \ref{thm:margin_generalization_uniform} to any strongly convex set. When combined with bounds on the multivariate Rademacher complexity, such as those developed in Section \ref{sec:ambuj_bounds}, we obtain generalization bounds with significantly better dependence on the number of parameters than Corollary \ref{cor:balls}.  
As a basic example, we also recover known results in binary classification, which is described in Example \ref{example:binary} below.

\begin{example}\label{example:binary}
In \cite{elmachtoub2017smart}, it is shown that the SPO loss corresponds exactly to the 0-1 loss in binary classification when $d = 1$, $S = [-1/2, +1/2]$, and $\calC = \{-1, +1\}$. In this case, using our notation, the margin value of a prediction $\hat{c}$ is $c\hat{c}$. 
It is also easily seen that $\ldiam(\calC) = \rho_2(\calC) = 1$, the $\gamma$-margin SPO loss corresponds exactly to the margin loss (or ramp loss) that interpolates between 1 and 0 when $c\hat{c} \in [0, \gamma]$
Furthermore, note that the interval $S = [-\tfrac{1}{2}, +\tfrac{1}{2}]$ is $2$-strongly convex \cite{garber2015faster}. In this case, the right hand side of the generalization bound of Theorem \ref{thm:margin_generalization}, for example, can be upper bounded by $O(\Rad(\calH)/\gamma + \sqrt{\log(1/\delta)/n})$, where the $O(\cdot)$ notation hides absolute constants. This recovers a well known result in binary classification \citep{koltchinskii2002empirical} and, altogether, Theorems \ref{thm:margin_generalization}, \ref{thm:margin_generalization_uniform}, and \ref{thm:strongly} generalize the well-known results on margin guarantees based on Rademacher complexity for binary classification. 
\Halmos \end{example}


\subsection{Polyhedral Sets}
Let us now assume that $S$ is polyhedral with a known convex hull representation, i.e., $S = \conv\{v_1, \ldots, v_K\}$ where $v_1, \ldots, v_K \in \bbR^d$ are given vectors.
We allow redundancies in the convex hull representation (i.e., it may be possible to express $v_j$ as a convex combination of other points in the basis), but for simplicity and ease of exposition we do require that $v_i \neq v_j$ for $i \neq j$.

Let $\calK_j := -N_S(v_j)$ be the negative normal cone at $v_j$ for $j = 1, \ldots, K$. Equivalently, we have that 
\begin{equation}\label{eq:poly-normals}
\calK_j = \{\hat{c} \in \bbR^d : \hat{c}^T(w - v_j) \geq 0 \text{ for all } w \in S\} = \{\hat{c} \in \bbR^d : \hat{c}^T(v_i - v_j) \geq 0 \text{ for all } i = 1, \ldots, K\} \ ,
\end{equation}
and hence $\calK_j$ is a polyhedral cone. Note also that, by the fundamental theorem of linear optimization, we have that $\cup_{j = 1}^K \calK_j = \bbR^d$. Moreover, $\calK_1, \ldots, \calK_K$ form part of a polyhedral complex called the normal fan of a polytope \citep{ziegler2012lectures}. In our context, we can use $\calK_1, \ldots, \calK_K$ to characterize the set $\calC^{\circ}$ and subsequently compute the distance to degeneracy function $\nu_S(\cdot)$. The following proposition exactly characterizes $\calC^{\circ}$ in terms of $\calK_1, \ldots, \calK_K$.
\begin{proposition}\label{prop:poly_circ}
Suppose that $S$ is polyhedral with $S = \conv\{v_1, \ldots, v_K\}$. Then, for any $\hat{c} \in \bbR^d$, $P(\hat{c})$ has a unique optimal solution if and only if there exists $j \in \{1, \ldots, K\}$ such that $\hat{c} \in \Int(\calK_j)$. Subsequently, it holds that $\calC^{\circ} = \bbR^d \setminus \cup_{j = 1}^K \Int(\calK_j)$.
\end{proposition}
\proof{Proof.}
If $P(\hat{c})$ has a unique optimal solution, then $w^\ast(\hat{c}) = v_j$ is the unique optimal solution for some $j \in \{1, \ldots, K\}$. Furthermore, we have that $\hat{c}^Tv_j < \hat{c}^Tv_i$ for all $i \in \{1, \ldots, K\}$, $i \neq j$, which implies that $\hat{c} \in \Int(\calK_j)$. Conversely, if $\hat{c} \in \Int(\calK_j)$ then \eqref{eq:poly-normals} implies that $\hat{c}^T(v_i - v_j) > 0$ for all $i \neq j$. Therefore, since any $w \in S$ is a convex combination of $v_1, \ldots, v_K$, it must be the case that $v_j$ is the unique optimal solution of $P(\hat{c})$. Note that we have shown that $\hat{c}$ is in the complement of $\calC^{\circ}$ if and only if $\hat{c} \in \cup_{j = 1}^K \Int(\calK_j)$, hence $\calC^{\circ} = \bbR^d \setminus \cup_{j = 1}^K \Int(\calK_j)$.
\Halmos \endproof

We are now ready to state our main theorem in the polyhedral case, which gives an exact formula for the distance to degeneracy function $\nu_S(\cdot)$ in terms of $v_1, \ldots, v_K$ and shows that the strength property is satisfied. Recall that the diameter of $S$ in the given norm is defined by $\diam(S) := \sup_{w_1, w_2 \in S}\|w_1 - w_2\|$, which in the polyhedral case satisfies $\diam(S) = \max_{i,j \in \{1, \ldots, K\}}\|v_i - v_j\|$.
\begin{theorem} \label{thm:poly}
Suppose that $S$ is not a singleton and is polyhedral with $S = \conv\{v_1, \ldots, v_K\}$. Then the distance to degeneracy function satisfies
\begin{equation}\label{dist_poly}
\nu_S(\hat{c}) = \min_{j: v_j \neq w^\ast(\hat{c})}\left\{\frac{\hat{c}^T(v_j - w^\ast(\hat{c}))}{\|v_j - w^\ast(\hat{c})\|}\right\} \ \text{ for all } \hat{c} \in \bbR^d \ ,
\end{equation}
and the strength property in \eqref{strength_property} is satisfied with $\mu \gets \frac{2}{\diam(S)}$.
\end{theorem}
\proof{Proof.}
Let us first consider the case where $\hat{c} \in \calC^{\circ}$, and hence $\nu_S(\hat{c}) = 0$. Then, since there are multiple optimal solutions of $P(\hat{c})$, there exists $v_j$ that is an optimal solution of $P(\hat{c})$ with $v_j \neq w^\ast(\hat{c})$. Thus, the right side of \eqref{dist_poly} is equal to zero, which matches $\nu_S(\hat{c})$.

Let us now consider the case where $\hat{c} \not\in \calC^{\circ}$.
Since $\nu_S(\cdot)$ is a distance function in $\bbR^d$, we have for any $\hat{c} \not\in \calC^{\circ}$ that $\nu_S(\hat{c}) = \dist_{\calC^{\circ}}^\ast(\hat{c}) = \sup_{r}\left\{r : B_\ast(\hat{c}, r) \subseteq \bbR^d \setminus \calC^{\circ}\right\}$. Thus, by Proposition \ref{prop:poly_circ}, we have that $\nu_S(\hat{c}) = \sup_{r}\left\{r : B_\ast(\hat{c}, r) \subseteq \cup_{j = 1}^K \Int(\calK_j)\right\}$ for $\hat{c} \not\in \calC^{\circ}$.

In the remainder of the proof, we let $j^\ast(\cdot) : \bbR^d \to \{1, \ldots, K\}$ be any mapping such that $v_{j^\ast(\hat{c})}$ is an optimal solution of $P(\hat{c})$ for all $\hat{c} \in \bbR^d$.
Now, for any $r > 0$ and $\hat{c} \not\in \calC^{\circ}$, using $\nu_S(\hat{c}) = \sup_{r}\left\{r : B_\ast(\hat{c}, r) \subseteq \cup_{j = 1}^K \Int(\calK_j)\right\}$ we have the following chain of equivalences:
\begin{align*}
\nu_S(\hat{c}) \geq r \ &\Longleftrightarrow
B_\ast(\hat{c}, r - \epsilon) \subseteq \cup_{j = 1}^K \Int(\calK_j) \text{ for all } \epsilon \in (0, r] \\ 
&\Longleftrightarrow B_\ast(\hat{c}, r - \epsilon) \subseteq \Int(\calK_{j^\ast(\hat{c})}) \text{ for all } \epsilon \in (0, r] \\
&\Longleftrightarrow B_\ast(\hat{c}, r) \subseteq \calK_{j^\ast(\hat{c})} \\
&\Longleftrightarrow (\hat{c} + \Delta)^T(v_j - v_{j^\ast(\hat{c})}) \geq 0 \text{ for all } j = 1, \ldots, K \text{ and } \Delta \text{ s.t. } \|\Delta\|_\ast \leq r \\
&\Longleftrightarrow \hat{c}^T(v_j - v_{j^\ast(\hat{c})}) - r\|v_j - v_{j^\ast(\hat{c})}\| \geq 0 \text{ for all } j = 1, \ldots, K \\
&\Longleftrightarrow r \leq \min_{j \neq j^\ast(\hat{c})}\left\{\frac{\hat{c}^T(v_j - v_{j^\ast(\hat{c})})}{\|v_j - v_{j^\ast(\hat{c})}\|}\right\} = \min_{j: v_j \neq w^\ast(\hat{c})}\left\{\frac{\hat{c}^T(v_j - w^\ast(\hat{c}))}{\|v_j - w^\ast(\hat{c})\|}\right\} \ .
\end{align*}
where the second equivalence holds since $\cup_{j = 1}^K \Int(\calK_j)$ is a disjoint union of open sets, the third equivalence holds since $\calK_{j^\ast(\hat{c})}$ is a closed set, the fifth equivalence uses the fact that $\|\cdot\|$ is the dual norm of $\|\cdot\|_\ast$, and the equality at the end uses the fact that $v_{j^\ast(\hat{c})} = w^\ast(\hat{c})$ since $\hat{c} \not\in \calC^{\circ}$. Since $\calC^{\circ}$ is a closed set by Proposition \ref{prop:poly_circ}, we have that $\nu_S(\hat{c}) > 0$. Furthermore, since $w^\ast(\hat{c})$ is the unique optimal solution of $P(\hat{c})$, we have that $\min_{j: v_j \neq w^\ast(\hat{c})}\left\{\frac{\hat{c}^T(v_j - w^\ast(\hat{c}))}{\|v_j - w^\ast(\hat{c})\|}\right\} > 0$. Therefore, applying the above chain of equivalences twice with both $r \gets \nu_S(\hat{c})$ and $r \gets \min_{j: v_j \neq w^\ast(\hat{c})}\left\{\frac{\hat{c}^T(v_j - w^\ast(\hat{c}))}{\|v_j - w^\ast(\hat{c})\|}\right\}$ yields \eqref{dist_poly}.

Now, to see that \eqref{strength_property} (strength property) holds, let $\hat{c} \in \bbR^d$ and $w \in S$ be arbitrary. Then, there exist nonnegative scalars $\alpha_1, \ldots, \alpha_K$ such that $w = \sum_{j = 1}^K \alpha_jv_j$ and $\sum_{j = 1}^K \alpha_j = 1$. By \eqref{dist_poly}, it holds that
\begin{equation}\label{dist_poly2}
\hat{c}^T(v_j - w^\ast(\hat{c})) ~\geq~ \nu_S(\hat{c})\|v_j - w^\ast(\hat{c})\| \ \text{ for all } j = 1, \ldots, K \ .
\end{equation}
Now we have
\begin{align*}
\hat{c}^T(w - w^\ast(\hat{c})) ~&=~ \sum_{j = 1}^K \alpha_j\hat{c}^T(v_j - w^\ast(\hat{c})) \\
~&\geq~ \nu_S(\hat{c})\sum_{j = 1}^K\alpha_j\|v_j - w^\ast(\hat{c})\| \\
~&\geq~ \nu_S(\hat{c})\|w - w^\ast(\hat{c})\| \\
~&\geq~ \left(\frac{\nu_S(\hat{c})}{\diam(S)}\right)\|w - w^\ast(\hat{c})\|^2 \ ,
\end{align*}
where the first inequality uses \eqref{dist_poly2}, the second uses convexity, and the third uses the definition of the diameter $\diam(S)$.
\Halmos \endproof

With Theorem \ref{thm:poly} in place, we can now apply the margin-based generalization bounds of Theorems \ref{thm:margin_generalization} and \ref{thm:margin_generalization_uniform} to any polyhedral set with a known convex hull representation. The most basic example of such a set is the unit simplex, which corresponds to the case of multiclass-classification as described in Examples \ref{ex:spo} and \ref{example:multi_part2}.
More generally, when combined with bounds on the multivariate Rademacher complexity, such as those developed in Section \ref{sec:ambuj_bounds}, we obtain generalization bounds in the polyhedral case that may have significantly better dependence on the number of parameters than Corollary \ref{cor:linear}.

\begin{example}[Continuation of Example \ref{ex:spo}]\label{example:multi_part2}
Recall that $S := \{w \in \bbR^d : \sum_{j = 1}^d w_j = 1, w \geq 0\}$ and, using the $\ell_2$-norm, we have that $\Delta(S) = 2$. Additionally, $\calC=\{-e_i| i=1,\ldots,d\}$ and therefore $\rho_2(\calC) = \ldiam(\calC) = 1$. In this case, the right hand side of the generalization bound of Theorem \ref{thm:margin_generalization}, for example, can be upper bounded by $O(\Rad(\calH)/\gamma + \sqrt{\log(1/\delta)/n})$, which recovers a well known result in multi-class classification (see, e.g., \cite{NIPS2018_7431} and the references therein). Altogether, Theorems \ref{thm:margin_generalization}, \ref{thm:margin_generalization_uniform}, and \ref{thm:strongly} generalize known results on margin guarantees based on Rademacher complexity for multi-class classification. 
\Halmos \end{example}

\section{Conclusions and Future Directions}

Our work extends learning theory, as developed for binary and multiclass classification, to predict-then-optimize problems in two very significant directions: (i) obtaining worst-case generalization bounds using combinatorial parameters that measure the capacity of function classes, and (ii) exploiting special structure in data by deriving margin-based generalization bounds that scale more gracefully w.r.t.\ problem dimensions. It also motivates several interesting avenues for future work. Beyond the margin theory, other aspects of the problem that lead to improvements over worst case rates should be studied. In this respect, developing a theory of local Rademacher complexity for predict-then-optimize problems would be a promising approach. 
Developing a theory of surrogate losses, especially convex ones, that are calibrated w.r.t.  the non-convex SPO loss will also be extremely important. Finally, the assumption that the optimization objective is linear could be relaxed to include non-linear objectives, in which case a new framework and methodology would be required. 

\subsubsection*{Acknowledgments}

OE thanks Rayens Capital for their support. AE acknowledges the support of NSF via grant CMMI-1763000. PG acknowledges the support of NSF Awards CCF-1755705 and CMMI-1762744. AT acknowledges the support of NSF via CAREER grant IIS-1452099 and of a Sloan Research Fellowship.

\bibliographystyle{ormsv080} 
\bibliography{new_references} 


\begin{APPENDIX}{}

\section{Proofs for Section \ref{sec:framework}}

\subsection{Proof of Corollary \ref{SPO_rademacher_cor}} \label{cor:SPO_rademacher_cor}   
\proof{Proof.} Let $f^* \in \argmin_{f \in \calH} \riskT(f)$. Then, 
\begin{align*}
\riskT(\hat{f}_n) -   \riskT(f^*) &= \riskT(\hat{f}_n)  - \empRiskT(\hat{f}_n) + \empRiskT(\hat{f}_n) - \empRiskT(f^*)  + \empRiskT(f^*) -   \riskT(f^*) \\
&\leq \riskT(\hat{f}_n)  - \empRiskT(\hat{f}_n)   + \empRiskT(f^*) -   \riskT(f^*)  
\end{align*}
The inequality follows from the fact that $\hat{f}_n$ minimizes the empirical risk, i.e., $\hat{f}_n \in \argmin_{f\in \calH} \empRiskT(f)$. From Theorem \ref{SPO_rademacher}, we know that
\begin{align*}
\riskT(\hat{f}_n)  - \empRiskT(\hat{f}_n) \leq 2\RadSPO(\calH) + \ldiam(\calC)\sqrt{\frac{\log(2/\delta)}{2n}} 
\end{align*}
with probability at least $1-\frac{\delta}{2}$. From From Hoeffding's inequality, we have that
\begin{align*}
\empRiskT(f^*) - \riskT(f^*) \leq \ldiam(\calC)\sqrt{\frac{\log(2/\delta)}{2n}}
\end{align*}
with probability at least $1-\frac{\delta}{2}$. Thus, with probability at least $1-\delta$ we can combine the three previous inequalities and obtain the desired result.
\Halmos \endproof

\section{Proofs for Section \ref{sec:polyhedron}}
\subsection{Proof of Theorem \ref{Main_Theo}} \label{sec:proofthm2}
\proof{Proof.} The proof is along the lines of Corollary 3.8 in \cite{mohri2018foundations}. Fix a sample of data $\mathcal{S}_n=(\mathbb{X},\mathbb{C})\in (\mathcal{X}, \mathcal{C})^n$, where $\mathbb{X}=(x_1,\ldots,x_n)$ and $\mathbb{C}=(c_1,\ldots, c_n)$. Let  $\mathfrak{F}_{|\mathbb{X}}:=\{(w^*(f(x_1)),\dots,w^*(f(x_n))): f \in \mathcal{H}\}$. 
From the definition of empirical Rademacher complexity, we have that
\begin{align*}
\RadHatSPO(\calH) &=  \bbE_{\sigma}\left[\sup_{f \in \calH}\frac{1}{n}\sum_{i = 1}^n \sigma_{i}\ellT(f(x_i), c_i)\right] \\
&=  \bbE_{\sigma}\left[\sup_{f \in \calH}\frac{1}{n}\sum_{i = 1}^n \sigma_{i} c_i^T(w^*(f(x_i))-w^*(c_i))\right]  \\
&=  \bbE_{\sigma}\left[\sup_{(w_1,\ldots,w_n) \in \mathfrak{F}_{|\mathbb{X}}}\frac{1}{n}\sum_{i = 1}^n \sigma_{i} c_i^T(w_i-w^*(c_i))\right]  \\
&\leq \ldiam(\calC) \sqrt{\frac{2 \log |\mathfrak{F}_{|\mathbb{X}}| }{n}}\\
&\leq \ldiam(\calC) \sqrt{\frac{2 d_N(w^*\mathcal{(H)}) \log (n |\mathfrak{S}|^2)}{n}}
\end{align*}
where the first inequality is directly due to Massart's lemma and the definition of $\ldiam(\calC)$ and the second inequality follows from the Natarajan Lemma (see Lemma 29.4 in \cite{shalev2014understanding}). When using the Natarajan Lemma, we observe that there $w^*(\cdot)$  has  at most $|\mathfrak{S}|$ possible values w.l.o.g. by the equivalence of optimizing over $S$ to optimizing over $\mathfrak{S}$. The bound for the expected version of the Rademacher complexity follows immediately from the bound on the empirical Rademacher complexity. Applying this bound with Theorem \ref{SPO_rademacher} concludes the proof.
\Halmos \endproof

\subsection{Proof of Corollary \ref{cor:linear}} \label{sec:cor1proof}
\proof{Proof.}
We will prove that $w^*(\calH_{\mathrm{lin}})$ is an instance of a linear multiclass predictor for a particular class-sensitive feature mapping $\Psi$. 
In our application of linear multiclass predictors, let $\Psi: \calX \times S \mapsto \mathbb{R}^{d \times p}$ be a function that takes as input a feature vector and an extreme point and outputs a matrix and let 
\begin{align*}
\calH_{\Psi}=\{ x  \mapsto \underset{w \in S}{\text{argmax}} \langle B, \Psi(x, w) \rangle : B \in \mathbb{R}^{d\times p} \}.
\end{align*}
We will show that, for $\Psi(x,w) = w x^T$, we have that $w^*(\calH_{\mathrm{lin}}) \subseteq \calH_{\Psi}$. Consider any  $f \in \calH_{\mathrm{lin}}$ and the associated matrix $B_f$. Then
\begin{align*}
w^*(B_f x)&\in \underset{w \in S}{\text{argmin}} (B_fx)^T w \\ 
&=  \underset{w \in S}{\text{argmax}} -\tr{(B_fx)^T w} \\
&=  \underset{w \in S}{\text{argmax}} -\tr{B_f^Tw x^T} \\
&=  \underset{w \in S}{\text{argmax}} \langle-B_f, w x^T\rangle .
\end{align*}
Thus, it is clear that for $\Psi(x,w)=w x^T$, choosing the function in $\calH_{\Psi}$ corresponding to $-B_f$ yields exactly the function $f$. Therefore  $w^*(\calH_{\mathrm{lin}}) \subseteq \calH_{\Psi}$.
Theorem 7 in \cite{daniely2014optimal} shows that $d_N(\calH_{\Psi}) \leq dp$. Since $w^*(\calH_{\mathrm{lin}}) \subseteq \calH_{\Psi}$, then  $d_N(w^*(\calH_{\mathrm{lin}})) \leq dp$. Combining this bound on the Natarajan dimension with Theorem \ref{Main_Theo} concludes the proof.
\Halmos \endproof

\subsection{Proof of Corollary \ref{cor:balls}} \label{sec:cor2proof}

\proof{Proof.}
Consider the smallest cardinality $\epsilon$-covering of the feasible region $S$ by Euclidean balls of radius $\epsilon$. From Example 27.1 in \cite{shalev2014understanding}, the number of balls needed is at most $\left( \frac{2\rho_2(S) \sqrt{d}}{\epsilon} \right)^d$. Let the set $\tilde{\mathfrak{S}}$ denote the centers of the balls from the smallest cardinality covering. Then it immediately follows that 
\begin{align} \label{eq:balls}
|\tilde{\mathfrak{S}}| \leq \left( \frac{2\rho_2(S) \sqrt{d}}{\epsilon} \right)^d.
\end{align}
Finally, let the function $\tilde{w}: S \mapsto \{1,\ldots, |\tilde{\mathfrak{S}}|\}$ be the function that takes a feasible solution in $S$ and maps it to the closest point in in $\tilde{\mathfrak{S}}$.

We can bound the empirical Rademacher complexity by
\begin{align}
\RadHatSPO(\calH) &=  \bbE_{\sigma}\left[\sup_{f \in \calH}\frac{1}{n}\sum_{i = 1}^n \sigma_{i}\ellT(f(x_i), c_i)\right]  \nonumber \\
&=  \bbE_{\sigma}\left[\sup_{f \in \calH}\frac{1}{n}\sum_{i = 1}^n \sigma_{i} c_i^T(w^*(f(x_i))-w^*(c_i))\right]   \nonumber \\
&=  \bbE_{\sigma}\left[\sup_{f \in \calH}\frac{1}{n}\sum_{i = 1}^n \sigma_{i}  c_i^T\left[ w^*(f(x_i))-\tilde{w}(w^*(f(x_i)))+\tilde{w}(w^*(f(x_i)))-w^*(c_i) \right] \right]   \nonumber \\
&\leq \bbE_{\sigma}\left[\sup_{f \in \calH}\frac{1}{n}\sum_{i = 1}^n \sigma_{i}  c_i^T\left[ w^*(f(x_i))-\tilde{w}(w^*(f(x_i)))\right] \right]   +  \bbE_{\sigma}\left[\sup_{f \in \calH}\frac{1}{n}\sum_{i = 1}^n \sigma_{i}  c_i^T\left[\tilde{w}(w^*(f(x_i)))-w^*(c_i) \right] \right]   \nonumber \\
&\leq 2\epsilon \rho_2(\calC)   +  \bbE_{\sigma}\left[\sup_{f \in \calH}\frac{1}{n}\sum_{i = 1}^n \sigma_{i}  c_i^T \left[\tilde{w}(w^*(f(x_i)))-w^*(c_i) \right] \right]   \nonumber \\
&\leq 2\epsilon \rho_2(\calC)   +  (\ldiam(\calC) + 2\epsilon \rho_2(\calC)) \sqrt{\frac{2 d_N(\tilde{w}(w^*\mathcal{(H)})) \log (n |\tilde{\mathfrak{S}}|^2)}{n}}  \label{eq:wtilde}
\end{align}
The first inequality follows from the triangle inequality. The second inequality follows from the fact that $w^*(f(x_i))$ and $\tilde{w}(w^*(f(x_i)))$ are at most $2 \epsilon$ away by the definition of $\tilde{w}$. In the worst case, the difference is in the direction of $c_i$, and   $c_i^T\left[ w^*(f(x_i))-\tilde{w}(w^*(f(x_i))) \right] \leq 2\epsilon||c_i||\leq 2\epsilon\rho_2(\calC)$. The third inequality follows from the same exact argument as that in Theorem \ref{Main_Theo} with two additional observations. First, the maximum value of $c_i^T \left[\tilde{w}(w^*(f(x_i)))-w^*(c_i) \right] $  is  $\ldiam(\calC) + 2\epsilon \rho_2(\calC)$ using a similar reasoning as in the second inequality. Second, the number of possible values for $\tilde{w}(\cdot)$ is at most $|\tilde{\mathfrak{S}}|$ by definition.

Thus, all that remains is to bound $d_N(\tilde{w}(w^*\mathcal{(H)}))$. To do this, we first observe that $d_N(w^*\mathcal{(H)}) \leq dp$, where the proof follows exactly that of Corollary \ref{cor:linear}. Finally, we observe that 
\begin{align} \label{eq:dN}
d_N(\tilde{w}(w^*\mathcal{(H)})) \leq d_N(w^*\mathcal{(H)}) \leq dp
\end{align}
since $\tilde{w}$ is simply a deterministic function, and thus the number of dichotomies (labelings) that can be generated by $\tilde{w}(w^*\mathcal{(H)})$ is at most that of $w^*\mathcal{(H)}$.
 
Now setting $\epsilon=\frac{1}{n}$, and combining Eq. \eqref{eq:balls}, Eq. \eqref{eq:wtilde}, and Eq. \eqref{eq:dN} yields
\begin{align}
\RadHatSPO(\calH)
&\leq \frac{2\rho_2(\calC)}{n}\left(1+   \sqrt{\frac{2 dp \log (n (2n\rho_2(S) \sqrt{d})^{2d})}{n}} \right)    +  \ldiam(\calC) \sqrt{\frac{2 dp \log (n (2n\rho_2(S) \sqrt{d})^{2d})}{n}}  \nonumber \\
&\leq \frac{2\rho_2(\calC)}{n}\left(1+   2d\sqrt{\frac{2p \log (2n\rho_2(S) d)}{n}} \right)    +  2d\ldiam(\calC) \sqrt{\frac{2p \log ( 2n\rho_2(S) d)}{n}} .
\label{eq:RCboundS}
\end{align}
Finally, combining Eq. \eqref{eq:RCboundS} with Theorem \ref{SPO_rademacher} yields
\begin{align*}
\riskT(f) &\leq \empRiskT(f) + 4d\ldiam(\calC) \sqrt{\frac{2p \log ( 2n\rho_2(S) d)}{n}} + \ldiam(\calC)\sqrt{\frac{\log(1/\delta)}{2n}} + O\left(\frac{1}{n}\right).
\end{align*}

\Halmos \endproof

\section{Proofs for Section \ref{sec:margin}}\label{sec:uniform_proof}

\subsection{Proof of Theorem \ref{thm:margin_generalization_uniform}} \label{sec:thm5proof}
\proof{Proof.}
The argument here follows closely the proof of Theorem 5.9 of \cite{mohri2018foundations}. Define $\epsilon := \ldiam(\calC)\sqrt{\frac{\log(2/\delta)}{2n}}$ and two sequences $\{\gamma_k\}_{k = 1}^\infty$ and $\{\epsilon_k\}_{k = 1}^\infty$ by
\begin{equation*}
\epsilon_k := \epsilon + \ldiam(\calC)\sqrt{\frac{\log(k)}{n}} \ , \text{ and } \gamma_k := \frac{\bar\gamma}{2^k} \ , \text{ for } k \geq 1 \ .
\end{equation*}
Define the following events:
\begin{align*}
&A_k := \left\{\sup_{f \in \calH}\left\{\riskT(f) - \empRiskTmark(f) - \frac{2\sqrt{2}(\rho_2(\calC) + \mu\cdot\ldiam(\calC))\Rad(\calH)}{\gamma_k\mu} - \epsilon_k\right\} > 0\right\} \text{ for } k \geq 1 \ , \\ 
&\tilde{A} := \bigcup_{k = 1}^\infty A_k \ , \text{ and } \\
&\check{A} := \left\{\sup_{f \in \calH, \gamma \in (0, \bar{\gamma}]}\left\{\riskT(f) - \empRiskTmar(f) - \frac{4\sqrt{2}(\rho_2(\calC) + \mu\cdot\ldiam(\calC))\Rad(\calH)}{\gamma\mu} - \ldiam(\calC)\sqrt{\frac{\log(\log_2(2\bar\gamma/\gamma))}{n}} - \epsilon\right\} > 0\right\} \ .
\end{align*}
Let us first argue that $\check{A} \subseteq \tilde{A}$. Indeed, suppose that $\check{A}$ occurs. Then, there exists some $f \in \calH$ and some $\gamma \in (0, \bar{\gamma}]$ such that:
\begin{equation}\label{green_day}
\riskT(f) - \empRiskTmar(f) - \frac{4\sqrt{2}(\rho_2(\calC) + \mu\cdot\ldiam(\calC))\Rad(\calH)}{\gamma\mu} - \ldiam(\calC)\sqrt{\frac{\log(\log_2(2\bar\gamma/\gamma))}{n}} - \epsilon > 0 \ . 
\end{equation}
By definition of the sequence $\{\gamma_k\}$, there exists $k \geq 1$ such that $\gamma_k \leq \gamma \leq 2\gamma_k$. Thus, $\gamma_k \leq \gamma$ implies that $\empRiskTmark(f) \leq \empRiskTmar(f)$. Moreover, $\gamma \leq 2\gamma_k$ implies that $-1/\gamma_k \geq -2/\gamma$, $k \leq \log_2(2\bar\gamma/\gamma)$, and thus 
\begin{equation*}
\epsilon_k = \epsilon + \ldiam(\calC)\sqrt{\frac{\log(k)}{n}} \leq \epsilon + \ldiam(\calC)\sqrt{\frac{\log(\log_2(2\bar\gamma/\gamma))}{n}} \ .
\end{equation*}
Now, combining the previous inequalities together with \eqref{green_day} yields:
\begin{equation*}
\riskT(f) - \empRiskTmark(f) - \frac{2\sqrt{2}(\rho_2(\calC) + \mu\cdot\ldiam(\calC))\Rad(\calH)}{\gamma_k\mu} - \epsilon_k > 0 \ ,
\end{equation*}
which means that the event $A_k$ and correspondingly the event $\tilde{A}$ have occurred. 

Now, for each $k \geq 1$, we apply Theorem \ref{thm:margin_generalization} using $\gamma \gets \gamma_k$ and $\delta \gets \exp((-2n\epsilon_k^2)/\ldiam(\calC)^2)$, which yields $\bbP(A_k) \leq \exp((-2n\epsilon_k^2)/\ldiam(\calC)^2)$.
We now apply $\bbP(\check{A}) \leq \bbP(\tilde{A})$ and the union bound to obtain:
\begin{align*}
\bbP(\check{A}) &\leq \sum_{k = 1}^\infty \exp\left(-\frac{2n\epsilon_k^2}{\ldiam(\calC)^2}\right) \\
&= \sum_{k = 1}^\infty \exp\left(-2n\left(\sqrt{\frac{\log(2/\delta)}{2n}} + \sqrt{\frac{\log(k)}{n}}\right)^2\right) \\
&< \sum_{k = 1}^\infty \exp\left(-(\log(2/\delta) + 2\log(k))\right) \\
&= \frac{\delta}{2}\sum_{k = 1}^\infty\frac{1}{k^2} = \frac{\delta}{2}\cdot\frac{\pi^2}{6} < \delta \ .
\end{align*}
Thus, we have completed the proof.
\Halmos \endproof

\subsection{Proof of Theorem \ref{thm:linear_class}} \label{sec:thm6proof}
\proof{Proof.}
Define $\vsigma_i = (\sigma_{i1},\ldots,\sigma_{id})^T$ for $i = 1, \ldots, n$ whose components are i.i.d. Rademacher random variables. Then, we have
\begin{align*}
\Rad(\calH_\calB)
&= \bbE\left[ \sup_{B \in \calB} \frac{1}{n} \sum_{i=1}^n \vsigma_i^T B x_i \right] \\
&= \bbE\left[ \sup_{B \in \calB} \frac{1}{n} \sum_{i=1}^n \tr{\vsigma_i^T B x_i} \right] \\
&= \bbE\left[ \sup_{B \in \calB} \frac{1}{n} \sum_{i=1}^n \tr{B x_i \vsigma_i^T} \right] \\
&= \bbE\left[ \sup_{B \in \calB} \tr{B \left( \frac{1}{n} \sum_{i=1}^n \vsigma_i x_i^T \right)^T} \right] \\
&= \bbE\left[ \sup_{B \in \calB} \langle B, \frac{1}{n} \sum_{i=1}^n \vsigma_i x_i^T \rangle \right] .
\end{align*}
Note that the inner product between matrices $B, A \in \bbR^{d \times p}$ is defined as
$\langle B, A \rangle := \sum_{i,j} B_{ij} A_{ij} = \tr{BA^T}$.
Now fix $x_1,\ldots,x_n$ and note that, by our assumption, we have, for all $i$,
\[
\sup_{\vsigma \in \{\pm1\}^d} \| \vsigma x_i^T \| \le \Omega .
\]
Let $\Theta$ be the random matrix $\frac{1}{n} \sum_{i=1}^n \vsigma_i x_i^T$. Choose arbitrary $\lambda > 0$. By Fenchel's inequality,
\[
\langle B, \Theta \rangle \le \frac{F(B)}{\lambda} + \frac{F^*(\lambda \Theta)}{\lambda}.
\]
Since $F(B) \le \beta^2$ for all $B \in \calB$, we have
\[
\sup_{B \in \calB} \langle B, \Theta \rangle \le \frac{\beta^2}{\lambda} + \frac{F^*(\lambda \Theta)}{\lambda}.
\]
Taking expectations (w.r.t. $\sigma_{ij}$) gives
\[
\bbE[ \sup_{B \in \calB} \langle B, \Theta \rangle ]
\le \frac{\beta^2}{\lambda} + \frac{\bbE[F^*(\lambda \Theta)]}{\lambda}
\]
Now let $Z_i = \frac{\lambda}{n} \vsigma_i x_i^T$ so that $S_n := \sum_{i=1}^n Z_i = \lambda\Theta$. Note that $\| Z_i \| \le \frac{\lambda}{n} \Omega$. So the conditions of Lemma 4 in \cite{kakade2009complexity} are satisfied with $V^2 = \lambda^2 \Omega^2/n^2$. That lemma gives us
$\bbE[F^*(\lambda \Theta)] \le \lambda^2 \Omega^2 / (2\alpha n)$. Plugging this above, we have
\[
\bbE[ \sup_{B \in \calB} \langle B, \Theta \rangle ]
\le \frac{\beta^2}{\lambda} + \frac{\lambda \Omega^2}{2\alpha n} .
\]
Setting $\lambda = (\beta/\Omega)\sqrt{2\alpha n}$ gives
\[
\bbE[ \sup_{B \in \calB} \langle B, \Theta \rangle ]
\le \Omega \beta \sqrt{\frac{2}{\alpha n}} 
\]
which completes the proof.
\Halmos \endproof

\section{Proofs for Section \ref{sec:margin-special}}\label{sec:vial_proof}

\subsection{Proof of Proposition \ref{prop:strong-normal}} \label{sec:prop1proof}
\proof{Proof.}
It is clear that the set on the right-hand side of \eqref{eq:strong-normal} is a subset of $N_S(\bar{w})$. To see the reverse, let $c \in N_S(\bar{w})$ and $w \in S$ be given. 
Define $\hat{w}(\lambda) := \lambda w + (1 - \lambda)\bar{w}$ and $r(\lambda) := \left(\tfrac{\bar\mu}{2}\right)\lambda(1 - \lambda)\|w - \bar{w}\|^2$ for $\lambda \in [0,1]$. 
By the $\bar\mu$-strong convexity of $S$, we have that $B(\hat{w}(\lambda), r(\lambda)) \subseteq S$. Now notice that
\begin{equation*}
c^T(\tilde{w} - \bar{w}) \leq 0 \ \text{ for all } \tilde{w} \in B(\hat{w}(\lambda), r(\lambda)) \ ,
\end{equation*}
since $c \in N_S(\bar{w})$ and $B(\hat{w}(\lambda), r(\lambda)) \subseteq S$.
Clearly the above condition is equivalent to:
\begin{equation*}
c^T\bar{w} ~\geq~ \max_{\tilde{w} \in B(\hat{w}(\lambda), r(\lambda))}\left\{c^T\tilde{w}\right\} ~=~ c^T\hat{w}(\lambda) + r(\lambda)\|c\|_\ast \ ,
\end{equation*}
where the equality above follows from the definition of the dual norm $\|\cdot\|_\ast$. Rearranging the above and using $\hat{w}(\lambda) - \bar{w} = \lambda(w - \bar{w})$ as well as the definition of $r(\lambda)$ yields:
\begin{equation*}
\lambda c^T(w - \bar{w}) ~\leq~ -\left(\tfrac{\bar\mu}{2}\right)\lambda(1 - \lambda)\|c\|_\ast\|w - \bar{w}\|^2 \ \text{ for all } \lambda \in [0,1] \ .
\end{equation*}
Now suppose that $\lambda > 0$. Dividing the above by $\lambda$ yields:
\begin{equation*}
c^T(w - \bar{w}) ~\leq~ -\left(\tfrac{\bar\mu}{2}\right)(1 - \lambda)\|c\|_\ast\|w - \bar{w}\|^2 \ \text{ for all } \lambda \in (0,1] \ .
\end{equation*}
Taking the limit as $\lambda \to 0$ yields
\begin{equation*}
c^T(w - \bar{w}) ~\leq~ -\left(\tfrac{\bar\mu}{2}\right)\|c\|_\ast\|w - \bar{w}\|^2 \ ,
\end{equation*}
which completes the proof.
\Halmos \endproof

\end{APPENDIX}

\clearpage
\newpage

\pagestyle{plain}
\setlength{\footskip}{8ex} 
\pagenumbering{arabic} 
\setcounter{page}{1}
\renewcommand{\thefigure}{\arabic{figure}}
\renewcommand{\thetheorem}{\arabic{theorem}}
\renewcommand{\thelemma}{\arabic{lemma}}
\renewcommand{\thecorollary}{\arabic{corollary}}
\renewcommand{\thesection}{\arabic{section}}
\renewcommand{\theremark}{\arabic{remark}}

\setcounter{figure}{0}
\setcounter{theorem}{0}
\setcounter{lemma}{0}
\setcounter{corollary}{0}
\setcounter{section}{1}
\setcounter{subsection}{0}
\setcounter{subsection}{0}


\end{document}